\documentclass{article}
\usepackage[utf8]{inputenc}
\usepackage[english]{babel}

\usepackage[a4paper,margin = 1.16in]{geometry}

\usepackage{microtype}
\usepackage{graphicx}
\usepackage{booktabs} 

\usepackage[toc,page]{appendix}
\usepackage{amssymb}
\usepackage{amsmath}
\usepackage{amsthm}
\usepackage{tikz-cd}
\usepackage{bbm}
\usepackage[ruled,vlined]{algorithm2e}

\usepackage{hyperref}
\usepackage{chngcntr}
\usepackage{apptools}
\usepackage{float}

\usepackage[skip=0pt]{caption}
\usepackage{subcaption}

\theoremstyle{definition}
\newtheorem{definition}{Definition}
\newtheorem{proposition}{Proposition}

\newtheorem{assumption}{Assumption}

\newtheorem*{claim}{Claim}

\newcommand{\R}{\mathbb{R}}
\newcommand{\N}{\mathbb{N}}
\newcommand{\E}{\mathbb{E}}
\newcommand{\Pp}{\mathbb{P}}

\newcommand{\Rd}{\mathbb{R}^{d}}
\newcommand{\Rnn}{\mathbb{R}^{n}}
\newcommand{\Rnnn}{\mathbb{R}^{n\times n}}
\newcommand{\Rdd}{\mathbb{R}^{d\times d}}

\newcommand{\Rnd}{\mathbb{R}^{n\times d}}
\newcommand{\dg}{^{\dagger}}

\newcommand{\G}{\mathcal{G}}
\newcommand{\Lc}{\mathcal{L}}

\newcommand{\Wh}{\widehat{w}}

\newcommand{\bfc}{\Sigma}
\newcommand{\bfm}{\Phi}
\newcommand{\bfq}{Q}
\newcommand{\bfz}{S}
\newcommand{\bfa}{A}
\newcommand{\bfb}{B}
\newcommand{\bfcc}{C}
\newcommand{\bfd}{D}

\newcommand{\bfch}{\widehat{\Sigma}}
\newcommand{\Vl}{V_{\ell_2}}
\newcommand{\Bl}{B_{\ell_2}}

\newcommand{\Wl}{w_{\ell_2}}
\newcommand{\Wc}{w_{V}}
\newcommand{\chalf}{{\Sigma^{\frac{1}{2}}}}
\newcommand{\cmhalf}{{{\Sigma}^{-\frac{1}{2}}}}

\newcommand{\cemhalf}{{{\Sigma_e}^{-\frac{1}{2}}}}
\newcommand{\mhalf}{{\bfm^{\frac{1}{2}}}}

\newcommand{\bfX}{X}
\newcommand{\bfZ}{Z}

\DeclareMathOperator*{\argmin}{arg\,min}

\newcommand{\Nn}{\{1,\dots,n\}}

\usepackage{natbib}
\setcitestyle{authoryear, open = {(}, close = {)}}

\title{On Optimal Interpolation in Linear Regression} 

\author{Eduard Oravkin, Patrick Rebeschini} 
\date{University of Oxford}

\begin{document}

\maketitle

\begin{abstract}
\noindent Understanding when and why interpolating methods generalize well has recently been a topic of interest in statistical learning theory. However, systematically connecting interpolating methods to achievable notions of optimality has only received partial attention. In this paper, we investigate the question of what is the optimal way to interpolate in linear regression using functions that are linear in the response variable (as the case for the Bayes optimal estimator in ridge regression) and depend on the data, the population covariance of the data, the signal-to-noise ratio and the covariance of the prior for the signal, but do not depend on the value of the signal itself nor the noise vector in the training data. We provide a closed-form expression for the interpolator that achieves this notion of optimality and show that it can be derived as the limit of preconditioned gradient descent with a specific initialization. We identify a regime where the minimum-norm interpolator provably generalizes arbitrarily worse than the optimal response-linear achievable interpolator that we introduce, and validate with numerical experiments that the notion of optimality we consider can be achieved by interpolating methods that only use the training data as input in the case of an isotropic prior. Finally, we extend the notion of optimal response-linear interpolation to random features regression under a linear data-generating model that has been previously studied in the literature.
\end{abstract}

\section{Introduction}\label{section_introduction}

Establishing mathematical understanding for the good generalization properties of interpolating methods, i.e.\ methods that fit the training data perfectly, has attracted significant interest in recent years. Motivated by the quest to explain the generalization performance of neural networks which have zero training error, for example even on randomly corrupted data \citep{zhang2016understanding}, this area of research has established results for a variety of models. For instance, in kernel regression, \citet{just_interpolate} provide a data-dependent upper bound on the generalization performance of the minimum-norm interpolator. By analyzing the upper bound, they show that small generalization error of the minimum-norm interpolator occurs in a regime with favourable curvature of the kernel, particular decay of the eigenvalues of the kernel and data population covariance matrices and, importantly, in an overparametrized setting. In random features regression, \citet{mei2019generalization} showed that for large signal-to-noise ratio and in the limit of large overparametrization, the optimal regularization is zero, i.e.\ the optimal ridge regressor is an interpolator. \citet{pragya} characterized the precise high-dimensional asymptotic generalization of interpolating minimum-$\ell_1$-norm classifiers and boosting algorithms which maximize the $\ell_1$ margin. \citet{benign} isolated a setting of benign overfitting in linear regression, dependent on notions of effective rank of the population covariance matrix, in which the minimum-norm interpolator has small generalization error. Similarly, this regime of benign overfitting occurs with large overparametrization.\\

Linear models, in particular, provide a fundamental playground to understand interpolators. On the one hand, in overparametrized regimes, interpolators in linear models are seen to reproduce stylized phenomena observed in more general models. For example, the double descent phenomenon, which was first empirically observed in neural networks \citep{belkin2018reconciling_double_descent}, has also featured
in linear regression \citep{hastie2019surprises_double_descent}. On the other hand, neural networks are known to be well-approximated by linear models in some regimes. For example, with specific initialization and sufficient overparametrization, two-layer neural networks trained with gradient descent methods are well-approximated by a first-order Taylor expansion around their initialization \citep{lazy_training_chizat}. This linear approximation can be split into a random features component and a neural-tangent component. The random features model, a two-layer neural network with randomly initialized first layer which is fixed during training, shares similar generalization behavior with the full neural network \citep{statdl}, and as such, the random features model provides a natural stepping stone towards tackling a theoretical understanding of neural networks.\\

A major focus of the interpolation literature has so far been to theoretically study if and when interpolating methods based on classical techniques such as ridge regression and gradient descent can have optimal or near-optimal generalization \citep{statdl}. However the question of understanding which interpolators are best, and designing data-dependent schemes to implement them, seems to have received only partial attention. Work investigating which interpolators are optimal in linear regression includes \citet{Harmless}, where the authors constructed the best-possible interpolator, i.e.\ a theoretical device which uses knowledge of the true parameter and training noise vector to establish a fundamental limit on how well {\it any} interpolator in linear regression can generalize. When the whitened features are sub-Gaussian, this fundamental limit is lower bounded by a term proportional to $n/d$, up to an additive constant and with high probability, which is small only in the regime of large overparametrization. Here, $n$ and $d$ are the size and the dimension of the data. While this interpolator provides the best-possible generalization error, the interpolator is not implementable in general, as one would need access to the realization of the true data-generating parameter $w^{\star}$ and the realization of the noise in the training data. \citet{rangamani2020interpolating} studied generalization of interpolators in linear regression and showed that the minimum-norm interpolator minimizes an {\it upper} bound on the generalization error related to stability. In \citep{mourtada2020exact}, it was shown that the minimum-norm interpolator is minimax optimal over any choice of the true parameter $w^{\star}\in\Rd$, distributions of the noise with mean $0$ and bounded variance, and for a fixed nondegenerate distribution of the features. \citet{amari2020does} computed the asymptotic risk of interpolating preconditioned gradient descent in linear regression and investigated the role of its implicit bias on generalization. In particular, they identified the preconditioning which leads to optimal asymptotic (as $d/n\to\gamma>1$ with $n,d\to\infty$) bias and variance, {\it separately}, among interpolators of the form $w = P\bfX^T(\bfX P\bfX^T)^{-1}y$ for some matrix $P$, where $X\in\Rnd$ is the data matrix, $y\in\R^{n}$ is the response vector. They showed that, within this class of interpolators, using the inverse of the population covariance matrix of the data as preconditioning achieves optimal asymptotic variance. However, the interpolator with optimal \emph{risk} is not given.\\

In this paper, we study the question of what is the optimal way to interpolate in overparametrized linear regression by procedures that do not use the realization of the true parameter generating the data, nor the realization of the training noise. The motivation for studying this question is twofold. First, in designing new ways to interpolate that are directly related to notions of optimality in linear models, we hope to provide a stepping stone to designing new ways to interpolate in more complex models, such as neural networks. Second, our results illustrate that there can be arbitrarily large differences in the generalization performance of interpolators, in particular considering the minimum-norm interpolator as a benchmark. This is a phenomenon that does not seem to have received close attention in the literature and may spark new interest in designing interpolators connected to notions of optimality.\\

We consider the family of interpolators that can be achieved as an arbitrary function $f$ of the data, population covariance, signal-to-noise ratio and the prior covariance such that $f$ is linear in the response variable $y$ (as the case for the Bayes optimal estimator in ridge regression). We call such interpolators {\it response-linear achievable} (see Definition \ref{def_response-linear_achievable_estimators}). We also introduce a natural notion of optimality that assumes that the realization of true data-generating parameter and the realization of the noise in the training data are unknown. Within this class of interpolators and under this notion of optimality, we theoretically compute the optimal interpolator and show that this interpolator is achieved as the implicit bias of a preconditioned  gradient descent with proper initialization. We refer to this interpolator as the {\it optimal response-linear achievable interpolator}.\\

Could it be that the commonly used minimum-norm interpolator is good enough so that the benefit of finding a better interpolator is negligible? We illustrate that the answer to this question is, in general, no. In particular, we construct an example in linear regression where the minimum-norm interpolator has arbitrarily worse generalization than the optimal response-linear achievable interpolator. Here, the variance (hence also generalization error) of the minimum-norm interpolator diverges to infinity as a function of the eigenvalues of the population covariance matrix, while the generalization error of the optimal response-linear achievable interpolator stays bounded, close to the optimal interpolator, i.e.\ the theoretical device of \citet{Harmless} which uses the true value of the signal and noise.\\

The optimal response-linear achievable interpolator uses knowledge of the population covariance matrix of the data (similarly as in \cite{amari2020does}), the signal-to-noise ratio, and the covariance of the true parameter (on which we place a prior distribution). Is it the case that the better performance of our interpolator is simply a consequence of this population knowledge? We provide numerical evidence that shows that the answer to this question is, in general, no. In particular, we construct an algorithm to approximate the optimal response-linear achievable interpolator which does not require any prior knowledge of the population covariance or the signal-to-noise ratio and uses only the training data $X$ and $y$, and we empirically observe that this new interpolator generalizes in a nearly identical way to the optimal response-linear achievable interpolator.\\

Finally, we show that the concept of optimal response-linear achievable interpolation can be extended to more complex models by providing analogous results for a random features model under the same linear data-generating regime as also considered in \citep{mei2019generalization}, for instance.

\section{Problem setup}\label{section_problem_setup}

\noindent In this paper we investigate overparametrized linear regression. We assume there exists $w^{\star}\in\Rd$ (unknown) so that $y_i = \langle w^{\star}, x_i\rangle + \xi_i$
for $i\in\Nn$, with i.i.d.\ noise $\xi_i\in \R$ (unknown) such that $\E(\xi_i) = 0, \E(\xi_i^2) = \sigma^2$ and i.i.d.\ features $x_i\in\Rd$ that follow a distribution ${\mathcal{P}_{x}}$ with mean $\E(x_i) = 0$ and covariance matrix $\E(x_ix_i^T) = \bfc$. We store the features in a random matrix $\bfX\in\Rnd$ with rows $x_i\in\Rd$, the response variable in a random vector $y\in\Rnn$ with entries $y_i\in\R$, and the noise in a random vector $\xi\in\Rnn$ with entries $\xi_i\in\R$. Throughout the paper we assume that $d\geq n$. We consider the whitened data matrix $\bfZ = \bfX\cmhalf \in \Rnd$, whose rows satisfy $ \E(z_iz_i^T) = I_d,$ where $I_d\in\Rdd$ is the identity matrix. We place a prior on the true parameter in the form $w^{\star} \sim 
{\mathcal{P}_{w^{\star}}}$ such that $\E(w^{\star}) = 0$ and  $\E(w^{\star}{w^{\star}}^T) = \frac{r^2}{d}\bfm$. Here, $\bfm$ is a positive definite matrix and $r^2$ is the signal. We sometimes abuse terminology and refer to $\bfm$ as the covariance of the prior even though $\frac{r^2}{d}\bfm$ is the covariance matrix. Our results will be proved in general, but for the sake of exposition it can be assumed that $x_i \sim \mathcal{N}(0,\bfc)$, $\mathbf{\xi} \sim \mathcal{N}(0,\sigma^2I_n)$ and $w^{\star} {\sim}\mathcal{N}(0,\frac{r^2}{d}\bfm)$. We also define the signal-to-noise ratio $\delta = r^2/\sigma^2$ and consider the squared error loss $\ell: (x,y) \in \R^{2} \mapsto (x - y)^2.$ Througout the paper, we assume the following two technical conditions hold.
\begin{assumption}\label{assumption_distr_x}
${\mathcal{P}_{x}}(x_i \in V)$ = 0 for any linear subspace $V$ of $\Rd$ with dimension smaller than $d$.
\end{assumption}
\begin{assumption}\label{assumption_distr_w_star}
For all Lebesgue measurable sets $A\subseteq\Rd$, $\nu(A)>0$ implies ${\mathcal{P}_{w^{\star}}}(w^{\star}\in A)>0$, where $\nu$ is the standard Lebesgue measure on $\Rd$.
\end{assumption}
Assumption \ref{assumption_distr_x} is needed only so that $\text{rank}(X) = n$ with probability $1$ (for a proof see \ref{appendix_ass_1_implies_rank_1}). A sufficient condition is that ${\mathcal{P}_{x}}$ has a density on $\Rd$. A sufficient condition for Assumption \ref{assumption_distr_w_star} is that ${\mathcal{P}_{w^{\star}}}$ has a positive density on $\Rd$. Now, our goal is to minimize the population risk
\begin{align*}
r(w) = \E_{\widetilde{x},\widetilde{\xi}} \big((\langle w, \widetilde{x}\rangle - \widetilde{y})^2\big), 
\end{align*}
or, equivalently, the excess risk $r(w) - r(w^{\star})$. Here, $(\widetilde{x},\widetilde{y}, \widetilde{\xi})$ is a random variable which follows the distribution of $(x_1,y_1,\xi_1),\dots, (x_n,y_n, \xi_n)$ and is independent from them. Throughout the paper we write $\mathbb{E}_z g(z,\tilde z)$ to denote the conditional expectation $\mathbb{E}( g(z,\tilde z) | \tilde z)$, for two random variables $z$ and $\tilde z$ and for a function $g$. The population risk satisfies
\begin{align}
r(w) &= (w-w^{\star})^T \Sigma (w-w^{\star}) + \sigma^2 = \Vert w-w^{\star} \Vert_{\bfc}^2 + r(w^{\star}), \label{expli_expli_form_of_risk}
\end{align}
where $\Vert w \Vert_{\bfc}^2 = w^T\bfc w$. For an estimator $w\in\Rd$, we define its bias and variance by the decomposition
\begin{align}
   \E_{\xi, w^{\star}}\; r(w) = B(w) + V(w), \label{eqn_bias_variance_decomposition}
\end{align}
where 
\begin{align}
   B(w) = \E_{\xi,w^{\star}}\Vert \E(w|w^{\star}, X) - w^{\star} \Vert^2_{\bfc} \qquad V(w) = \E_{\xi, w^{\star}}\Vert w - \E(w|w^{\star}, X) \Vert^2_{\bfc}. \label{eqn_def_bias_vriance}
\end{align}
One of the main paradigms to minimize the (unknown) population risk is based on minimizing the empirical risk $R(w) = \frac{1}{n} \sum^{n}_{i=1} (\langle w,x_i\rangle -  y_i)^{2} = \frac{1}{n} \sum^{n}_{i=1} \ell(w^Tx_i,y_i)$ \citep{vapnik}. In our setting, minimizing the empirical risk is equivalent to finding $w\in\Rd$ such that $\bfX w = y$.

\section{Interpolators}
An interpolator is any minimizer of the empirical risk. Let $\G$ be the set of interpolators, which in linear regression can be written as
\begin{align*}
\G = \{ w\in\Rd : \bfX w = y\}.    
\end{align*}
As $\text{rank}(\bfX) = n$ with probability $1$, we have $\G \ne \emptyset$ with probability $1$. In linear regression, the implicit bias of gradient descent initialized at $0$ is the minimum-norm interpolator \citep{optimization_geometry}. We define the minimum-norm interpolator by
\begin{align*}
    \Wl &= \argmin_{w\in\Rd \, : \, \bfX w = y} \Vert w \Vert_2^2 = \bfX\dg y, 
\end{align*}
where $\bfX\dg\in\R^{n\times d}$ is the Moore-Penrose pseudoinverse \citep{penrose_pseudoinv_def}. As $\text{rank}(\bfX) = n$, we can also write $\bfX\dg = \bfX^T(\bfX\bfX^T)^{-1}$. The second interpolator of interest is a purely theoretical device, previously used in \citep{Harmless} to specify a fundamental limit to how well any interpolator in linear regression can generalize.
\begin{definition}\label{best_possible_interpolator} The \emph{best possible interpolator} is defined as
\begin{align*}
W_{b} &= \argmin_{w\in\G} r(w). 
\end{align*}
\end{definition}
We can write
\begin{align*}
 W_{b} = \argmin_{w\in\Rd \, : \, \bfX w = y} \Vert \chalf(w-w^{\star}) \Vert_2^2, 
\end{align*}
and after a linear transformation and an application of a result on approximate solutions to linear equations \citep{penrose_lin_eqns}, we obtain
\begin{align}
     W_{b} = w^{\star} + \cmhalf (\bfX\cmhalf)^{\dagger} \xi. \label{explicit_form_of_best_parameter}
\end{align}
We notice that the best possible interpolator fits the signal perfectly by having access to the true parameter $w^{\star}$ and fits the noise through the term $\cmhalf (\bfX\cmhalf)^{\dagger} \xi$ by having access to the noise vector $\xi$ in the training data. In general, this interpolator cannot be implemented as it requires access to the unknown quantities $w^{\star}$ and $\xi$. We are interested in interpolators which can be achieved by some algorithm using the data $\bfX$ and $y$. 
\begin{definition}\label{def_achievable_estimator}
We define an estimator $w\in \Rd$ to be \emph{achievable} if there exists a function $f$ such that $w = f(\bfX, y, \bfc, \bfm, \delta)$.
\end{definition}

In our definition of achievability, we allow for knowledge of the population data covariance, the signal-to-noise ratio, and the prior covariance to define a fundamental limit to what generalization performance can be achieved also without access to these quantities, and we later empirically show that we can successfully approach this limit using only the knowledge of the training data $X$ and $y$, in considered examples (see Section \ref{section_approximation_of_the_true_covariance}). Moreover, our theory is also useful in situations when one has access to some prior information about the regression problem which they can incorporate into an estimate of $\bfc$, $\delta$, $\bfm$ (for example, one may know the components of $x_i$ are independent and hence $\bfc$ is diagonal) and hence it is relevant to consider a broader class than $w = f(X,y)$.

\begin{definition}\label{def_response-linear_achievable_estimators}
We define the set of \emph{response-linear achievable estimators} by 
\begin{align*}
\Lc = \{\:w\in \Rd : \exists f \text{ such that } w = f(\bfX, y, \bfc, \bfm, \delta) \text{ where } y\in\Rnn\mapsto f(\bfX, y, \bfc, \bfm, \delta)\text{ is linear}\:\}.
\end{align*}
\end{definition}
Linearity of $y\in\Rnn\mapsto f(\bfX, y, \bfc, \bfm, \delta)$ is equivalent to $f(\bfX,y,\bfc,\bfm,\delta) = g(\bfX,\bfc,\bfm,\delta)y$, where $g$ is any function which has image in $\mathbb{R}^{d\times n}$. The notion of optimality that we introduce is that of the optimal response-linear achievable interpolator, which is the interpolator that minimizes the expected risk in the class $\Lc$. 

\begin{definition}\label{def_optimal_response-linear_achievable_interpolator}
We define the \emph{optimal response-linear achievable interpolator} by
\begin{align}
   w_{O} = \argmin_{w\in\G\cap\Lc} \E_{\xi,w^{\star}} r(w) - r(w^{\star}).\label{eqn_def_bla}
\end{align}
\end{definition}

\section{Main results}
By definition, the interpolator $w_{O}$ has the smallest expected risk among all response-linear achievable interpolators. Our first contribution is the calculation of its exact form.
\begin{proposition}\label{prop_bla}
The optimal response-linear achievable interpolator satisfies
\begin{align}
    w_{O} &= \bigg(\frac{\delta}{d}\bfm\bfX^T \!+\! \cmhalf(\bfX\cmhalf)\dg\bigg)\bigg( I_n \!+\! \frac{\delta}{d}\bfX\bfm\bfX^T\bigg)^{-1}\!\!y.\label{eqn_bla_explicit_form}
\end{align}
\end{proposition}
For an isotropic prior $\bfm = I_d$, $w_{O}$ depends only on the population covariance $\bfc$ and the signal-to-noise ratio $\delta$ so that $w_{O}$ can be approximated using estimators of these quantities, which is what we do in Sections \ref{section_regimes_of} and \ref{appendix_empirical_comparison}. Even if $\bfm \ne I_d$, one might have some information about the prior covariance, which can be incorporated into an estimate ${\widehat{\Phi}}$ and used instead of $\Phi$. However, even if no such estimate is available, in Section \ref{subsection_non_isotropic} we empirically show that, in our examples, using ${\widehat{\Phi}} = I_d$ when $\Phi \ne I_d$ has a small effect on generalization. \\

Secondly, using results of \cite{optimization_geometry} on the implicit bias of converging mirror descent, we show that the optimal response-linear interpolator is the limit of gradient descent preconditioned by the inverse of the population covariance, provided that it converges and is suitably initialized.

\begin{proposition}\label{prop_bla_implicit_bias_of_preconditioned_GD}
The optimal response-linear achievable interpolator is the limit of preconditioned gradient descent
\begin{equation}
w_{t+1} = w_{t} - \eta_t{\bfc}^{-1}\nabla R(w_{t}), \label{eqn_preconditioned_gd_update_rule}
\end{equation}
provided that the algorithm converges, initialized at
\begin{align}
w_0 = \frac{\delta}{d}\bfm\bfX^{T}\bigg( I_n + \frac{\delta}{d}\bfX\bfm\bfX^T\bigg)^{-1}\! y. \label{eqn_initialization_of_bla}
\end{align}
\end{proposition}

The interpolator $w_{O}$ does not have the smallest bias or the smallest variance in the bias-variance decomposition $\E_{\xi, w^{\star}}\; r(w) = B(w) + V(w)$, but rather achieves a balance. This is related to the results of \citep{amari2020does}. Their setting looks at interpolators achieved as the limit of preconditioned gradient descent in linear regression (preconditioned with some matrix $P$) and initialized at $0$. Such interpolators can be written as $w = PX^T(X PX^T)^{-1}y$. For these interpolators, they compute the risk of $w$, separate the risk into a variance and a bias term and using random matrix theory they find what the variance and bias terms converge to when $d\to\infty, n \to\infty$ in a way such that $d/n\to\gamma > 1$. For these calculations to hold, they assume that the spectral distribution of $(\Sigma_d)_{d\in\mathbb{N}}$ converges weakly to a distribution supported on $[c, C]$ for some $c,C>0$. Then, after obtaining the limiting variance and bias, they prove which matrices $P$ minimize these limits \emph{separately} (not their sum, which is the overall asymptotic risk).\\

We approach the problem from the other direction. That is, we do not a priori consider interpolators that can be achieved as limits of specific algorithms, but we directly look at which interpolator minimizes the risk as a whole (not bias and variance separately). Only after computing the optimal response-linear interpolator, we show in Proposition \ref{prop_bla_implicit_bias_of_preconditioned_GD} that the interpolator \emph{is} in fact the limit of preconditioned gradient descent, however with a specific initialization. Our results hold for every finite $d\geq n$ and we do not put assumptions on the eigenvalues or the spectral distribution of $\Sigma$.\\

In particular, we can recover the results of \citep{amari2020does} as a special case of our Proposition \ref{prop_bla}. If we take the signal-to-noise ratio $\delta\to 0$ (by taking $r^2\to 0$) in Proposition 1, we obtain the matrix $P$ which achieves optimal variance and if we take $\delta\to \infty$ (by taking $\sigma^2\to 0$) in Proposition \ref{prop_bla}, we obtain the matrix $P$ which achieves optimal bias. Moreover, we provide a further extension in Proposition \ref{prop_preconditioned_GD_achieves_optimal_variance}.\\

We show that the preconditioned gradient descent $w_{t+1} = w_{t} - \eta_t{\bfc}^{-1}\nabla R(w_{t})$ achieves optimal variance among \emph{all} interpolators when initialized at \emph{any} deterministic $w_0$ and for \emph{any} finite $d,n\in \mathbb{N}$.
 
\begin{proposition}\label{prop_preconditioned_GD_achieves_optimal_variance}
The limit of preconditioned gradient descent  $w_{t+1} = w_{t} - \eta_t{\bfc}^{-1}\nabla R(w_{t})$
initialized at a deterministic $w_0\in\Rd$, provided that it converges, satisfies
\begin{align}
\lim_{t\to\infty}w_{t} = \argmin_{w\in\G} V(w) \label{eqn_pgd_optimal_variance}.
\end{align}
\end{proposition}
We note that the optimal variance is achieved among all interpolators, not only among response-linear achievable interpolators.\\

A natural question to ask is whether the optimal response-linear achievable interpolator $w_{O}$ provides a significant benefit compared to other interpolators. A second question is whether we can successfully approximate the optimal response-linear achievable interpolator without knowledge of the population covariance $\bfc$ and the signal-to-noise ratio $\delta$. In the following section, we illustrate that both the interpolator with optimal variance and the interpolator with optimal bias can generalize arbitrarily badly in comparison to $w_{O}$ as a function of the eigenvalues of the population covariance. In the same regimes where this happens, we present numerical evidence that we can successfully approximate $w_{O}$ by an empirical interpolator $w_{Oe}$ without any prior knowledge of $\bfc$ or $\delta$ by using the Graphical Lasso estimator \citep{glasso} of the covariance matrix $\bfc$ and choosing the empirical estimate of $\delta$ by crossvalidation on a subset of the data. 

\begin{section}{Comparison of interpolators}\label{section_regimes_of}
First, we present an example where the minimum-norm interpolator $\Wl$ generalizes arbitrarily worse than the best response-linear achievable interpolator $w_{O}$. Second, we give an example where an interpolator with optimal variance generalizes arbitrarily worse than $w_{O}$. This shows that arbitrarily large differences in test error are possible within the class of estimators which have zero training error.\\

The examples we consider take place in a setting where $x_i \sim \mathcal{N}(0,\bfc)$ and $w^{\star} \sim \mathcal{N}(0,\frac{r^2}{d}\Phi)$. Therefore, throughout Section \ref{section_regimes_of} we assume ${\mathcal{P}_{x}} = \mathcal{N}(0,\bfc)$ and ${\mathcal{P}_{w^{\star}}} = \mathcal{N}(0,\frac{r^2}{d}\Phi)$. Before presenting these examples, we discuss approximating $w_{O}$ by an interpolator, $w_{Oe}$, which uses only the data $\bfX$ and $y$. 

\begin{subsection}{Empirical approximation}\label{section_approximation_of_the_true_covariance}
The interpolator $w_{O}$ is the limit of the algorithm
\begin{align}
w_{t+1} = w_{t} - \eta_t{\bfc}^{-1}\nabla R(w_{t}). \label{eqn_preconditioned_gd_update_rule_again} 
\end{align} 
The population covariance $\bfc$ is required to run this algorithm. However, the matrix $\bfc$ is usually unknown in practice. One may want to estimate $\bfc$. However, if one replaces $\bfc$ by $\widetilde{\bfc} = \bfX^T\bfX/n + \lambda I_d$ (with $\lambda\geq 0$), then the limit of (\ref{eqn_preconditioned_gd_update_rule_again}) is the same as the limit of gradient descent (provided that both algorithms converge). This is because, using the singular value decomposition of $\bfX$, one can show
\[{\widetilde{\bfc}}^{-\frac{1}{2}} (\bfX{\widetilde{\bfc}}^{-\frac{1}{2}})^{\dagger}y = \bfX\dg y. \]
The preconditioned gradient update $w_{t+1} - w_{t} = \eta_tP^{-1}\nabla R(w_{t})$ has to \emph{not} belong to $\textrm{Im}(\bfX^T)$ in order to \emph{not} converge to the minimum-norm interpolator. Hence, using $P = \widetilde{\bfc} = \bfX^T\bfX/n + \lambda I_d$ (or, for example, also the Ledoit-Wolf shrinkage approximation \citep{ledoitwolf}) in preconditioned gradient descent removes the benefit of preconditioning in terms of generalization of the limit.\\ 

We use the Graphical Lasso approximation \citep{glasso}. We empirically observe that in the examples considered in this paper (Figures \ref{fig_variance_to_infty}, \ref{fig_bias_to_infty}, \ref{fig_autoregressive}, \ref{fig_exponential}, \ref{fig_nonisoregime_1}, \ref{fig_nonisoregime_2}) using the Graphical Lasso covariance $\bfc_{e}$ instead of $\bfc$ has nearly no effect on generalization. Under specific assumptions, \citet{ravikumar2008highdimensional} provide some convergence guarantees of the Graphical Lasso. \\

In regards to approximating the signal-to-noise ratio $\delta$, we choose $\delta_e$ that minimizes the crossvalidated error on random subsets of the data. In this way, we arrive at the interpolator 
\begin{align}
w_{Oe} & \!=\! \bigg(\frac{\delta_e}{d}\bfX^T \!+\! \cemhalf(\bfX\cemhalf)\dg\bigg)\bigg( I_n \!+\! \frac{\delta_e}{d}\bfX\bfX^T\bigg)^{-1}\!\!y,\label{eqn_wOe_definition}
\end{align}
which approximates $w_{O}$ and is a function of only $\bfX$ and $y$. We note that the interpolator $w_{Oe}$ uses $I_d$ in place of the prior covariance matrix. \\

In the experiments (Figures \ref{fig_variance_to_infty}, \ref{fig_bias_to_infty}, \ref{fig_autoregressive}, \ref{fig_exponential}, \ref{fig_nonisoregime_1}, \ref{fig_nonisoregime_2})  we used the Graphical  Lasso implementation of scikit-learn \citep{scikit-learn} with parameter $\alpha = 0.25$ ($\alpha$ can also be crossvalidated for even better performance) and in estimating $\delta$, for each $\delta_e$ in $\{0.1,0.2,\dots,1,2,\dots,10\}$, we computed the validation error on a random, unseen tenth of the data and averaged over $10$ times. The $\delta_e$ with smallest crossvalidated error was chosen.
\end{subsection}

\begin{subsection}{Random matrix theory concepts}\label{subsection_random_matrix}
For presenting the discussed examples we need to review some concepts from random matrix theory.
\begin{definition}\label{spectral_measure}
For a symmetric matrix $\bfc\in\Rdd$ with eigenvalues $\lambda_1\geq\lambda_2\geq\dots \geq \lambda_d\geq 0$ we define its spectral distribution by $\mathcal{F}_{\bfc}(x) = \frac{1}{d}\sum_{i=1}^d \mathbbm{1}_{[\lambda_i,\infty)}(x)$.
\end{definition}

The following assumptions will be occasionally considered for the covariance matrix $\Sigma$.
\begin{assumption}\label{ass_upper_boudn_on_eval}
There exists $k_{\text{max}}>0$ such that $\lambda_{\text{max}}(\bfc) \leq k_{\text{max}}$ uniformly for $d\in\N$.
\end{assumption}
\begin{assumption}\label{ass_lower_boudn_on_eval}
There exists $k_{\text{min}}>0$ such that $k_{\text{min}}\leq \lambda_{\text{min}}(\bfc)$ uniformly for $d\in\N$.
\end{assumption}
\begin{assumption}\label{ass_convergence_spectral_ditrib}
The spectral distribution $\mathcal{F}_{\bfc}$ of the covariance matrix $\bfc$ converges weakly to a distribution $\mathcal{H}$ supported on $[0,\infty)$. 
\end{assumption}
\citet{marchenko-pastur} showed that there exists a distribution $\widetilde{\mathcal{F}}_{\gamma}$ such that $\mathcal{F}_{\frac{\bfZ^T\bfZ}{n}} \longrightarrow \widetilde{\mathcal{F}}_{\gamma},$
weakly, with probability $1$ as $n\to \infty,d\to\infty$ with $d/n\to\gamma$. In our discussion, $x_i = \bfc^{\frac{1}{2}}z_i$, where $z_i{\sim}\mathcal{N}(0,I_d)$ independently. Then, under Assumption \ref{ass_convergence_spectral_ditrib}, it can be shown that the spectral distribution of $\bfch = \bfX^T\bfX/n = \bfc^{\frac{1}{2}}\bfZ^T\bfZ\bfc^{\frac{1}{2}}/n$
converges weakly, with probability $1$ to a distribution supported on $[0,\infty)$, which we denote by $F_{\gamma}$, see e.g. \citep{silvestrein_and_choi}. Similar arguments also show that the spectral distribution of $\bfX\bfX^T/n\in\Rnnn$ converges weakly, with probability $1$.\\

\begin{definition}\label{definition_Stieltjes_transform}
For a distribution $\mathcal{F}$ supported on $[0,\infty)$, we define the \emph{Stieltjes transform of $\mathcal{F}$}, for any $z\in\mathbb{C}\setminus\mathbb{R}^{+}$ by 
\begin{align*}
    m_{\mathcal{F}}(z) = \int_{0}^{\infty}\frac{1}{\lambda - z}d\mathcal{F}(\lambda).
\end{align*}
\end{definition}

The weak convergence of the spectral distribution of $\bfch$ to $\mathcal{F}_{\gamma}$ is equivalent to
$m_{\bfch}(z) \to m(z)$ and $m_{\bfX\bfX^T/n}(z) \to v(z)$
almost surely for all $z\in\mathbb{C}\setminus\mathbb{R}^{+}$, where $m$ and $v$ are the Stieltjes transforms of $\mathcal{F}_{\gamma}$ and the limiting spectral distribution of $\bfX\bfX^T/n$, respectively (see e.g. Proposition 2.2 of \citep{hachem2007}). We call $v$ the companion Stieltjes transform of $\mathcal{F}_{\gamma}$.
\end{subsection}

\begin{subsection}{\bf Diverging variance of interpolator 
with optimal bias}\label{subsectionn_variance_gd_to_infty}\label{section_diverging_bias}
Using $w_0 = 0$ in Proposition \ref{prop_preconditioned_GD_achieves_optimal_variance}, we choose the interpolator with optimal variance to be (see \ref{supplementary_material_proof_Proposition_1})
\begin{equation}
\Wc = \cmhalf(\bfX\cmhalf)\dg y.\label{eqn_def_best_variance}
\end{equation}
When $\bfm = I_d$, the interpolator with best bias among response-linear achievable interpolators is the minimum-norm interpolator (see Section \ref{appendix_interpolator_with_optimal_bias}). We identify an example where the minimum-norm interpolator $\Wl$ generalizes arbitrarily worse than the best response-linear achievable interpolator $w_{O}$. For this, we exploit results of \citet{hastie2019surprises_double_descent} on computing the asymptotic risk of the minimum-norm interpolator. They show that if $\bfm = I_d$, under Assumptions \ref{ass_upper_boudn_on_eval}, \ref{ass_lower_boudn_on_eval}, \ref{ass_convergence_spectral_ditrib} and if $\frac{d}{n}\to\gamma > 1$ with $n\to\infty,d\to\infty$ then with probability $1$,
\begin{align}
    \E_{\xi, w^{\star}} r(\Wl) - r(w^{\star}) \longrightarrow \frac{r^2}{\gamma v(0)} + \sigma^2\bigg( \frac{v'(0)} {v(0)^2} -1 \bigg), \label{eqn_asymptotic_risk_general_case}
\end{align}
where $v$ is the companion Stieltjes transform introduced in Section \ref{subsection_random_matrix}. In comparison, similarly as in \citep{amari2020does}, the asymptotic risk of the best variance estimator $\Wc$ satisfies
that under Assumption \ref{ass_upper_boudn_on_eval} and \ref{ass_convergence_spectral_ditrib}, if $n,d\to\infty$ with $\frac{d}{n}\to\gamma > 1$ then with probability $1$ we have that
\begin{align}
\lim_{d\to\infty}\E_{\xi, w^{*}} r(\Wc) - r(w^{\star}) = r^2\frac{\gamma-1}{\gamma}\int_{0}^{\infty}s\; d\mathcal{H}(s) + \frac{\sigma^2}{\gamma - 1}. \label{eqn_mirror_asymptotic_risk_bayesian_2}
\end{align}

An alternative way is to write $\int_{0}^{\infty}s \; d\mathcal{H}(s) = \lim_{d\to\infty} \text{Tr}(\bfc)$. This result follows by an application of Theorem 1 of \citep{rubio_mestre}, which is in the supplementary material for completeness.\\

Now we find a regime of covariance matrices $\bfc$, for which the variance term of the minimum-norm solution, $\Vl = \sigma^2(\frac{v'(0)}{v(0)^2}-1),$ diverges to infinity, while the risk of $w_{O}$ stays bounded and close to optimal. For this, we consider a generalization of the spike model of covariance matrices \citep{spike_model_silverstein, spike_johnstone}, which is a fundamental model in statistics. Here $\bfc = \text{diag}(\rho_1,\dots, \rho_1, \rho_2, \dots, \rho_2)\in\Rdd,$ where the number of $\rho_1$s is $d\cdot\psi_1$ with $\psi_1 \in [0,1]$. This model was also considered in \citep{dominic} where it is called the strong weak features model. In this regime, it is possible to explicitly calculate the companion Stieltjes transform $v(0)$ and $v'(0)$ of (\ref{eqn_asymptotic_risk_general_case}). In the case that $\gamma = 2, \psi_1 = 1/2$ we have
\begin{align}
\Vl = \sigma^2\bigg(\frac{v'(0)}{v(0)^2}-1\bigg) = \frac{\sigma^2}{2}\bigg(\sqrt{\frac{\rho_1}{\rho_2}} + \sqrt{\frac{\rho_2}{\rho_1}} + 2\bigg). \label{variance_gd_strong_weak}
\end{align}
If we fix $\rho_1 = 1$ and take $\rho_2 \to 0$, then the variance term $\Vl$ diverges to infinity. This also means that the asymptotic risk of the minimum-norm interpolator diverges to infinity. Moreover, the asymptotic risk of $\Wc$ in (\ref{eqn_mirror_asymptotic_risk_bayesian_2}) evaluates to
\begin{equation}
\lim_{d\to\infty}\E_{\xi, w^{*}} r(\Wc) - r(w^{\star}) = \bigg(\psi_1\rho_1 + (1-\psi_1)\rho_2\bigg)\bigg(1-\frac{1}{\gamma}\bigg) + \frac{\sigma^2}{\gamma - 1}. \label{auxi_hh}
\end{equation}
In addition, by construction of $w_{O}$, we know that $\E_{\xi,w^{\star}} r(w_{O}) \leq \E_{\xi,w^{\star}} r(\Wc)$
and therefore the asymptotic limit of $\E_{\xi,w^{\star}} r(w_{O}) - r(w^{\star})$, as $d/n\to\gamma>1$, stays bounded by (\ref{auxi_hh}) as $\rho_2 \to 0$. The expected generalization error in the setting described above is illustrated in Figure \ref{fig_variance_to_infty}.

\vspace{-0.3cm}
\begin{figure}[H]
    \begin{minipage}{0.49\textwidth}
        \centering
        \includegraphics[scale=0.45]{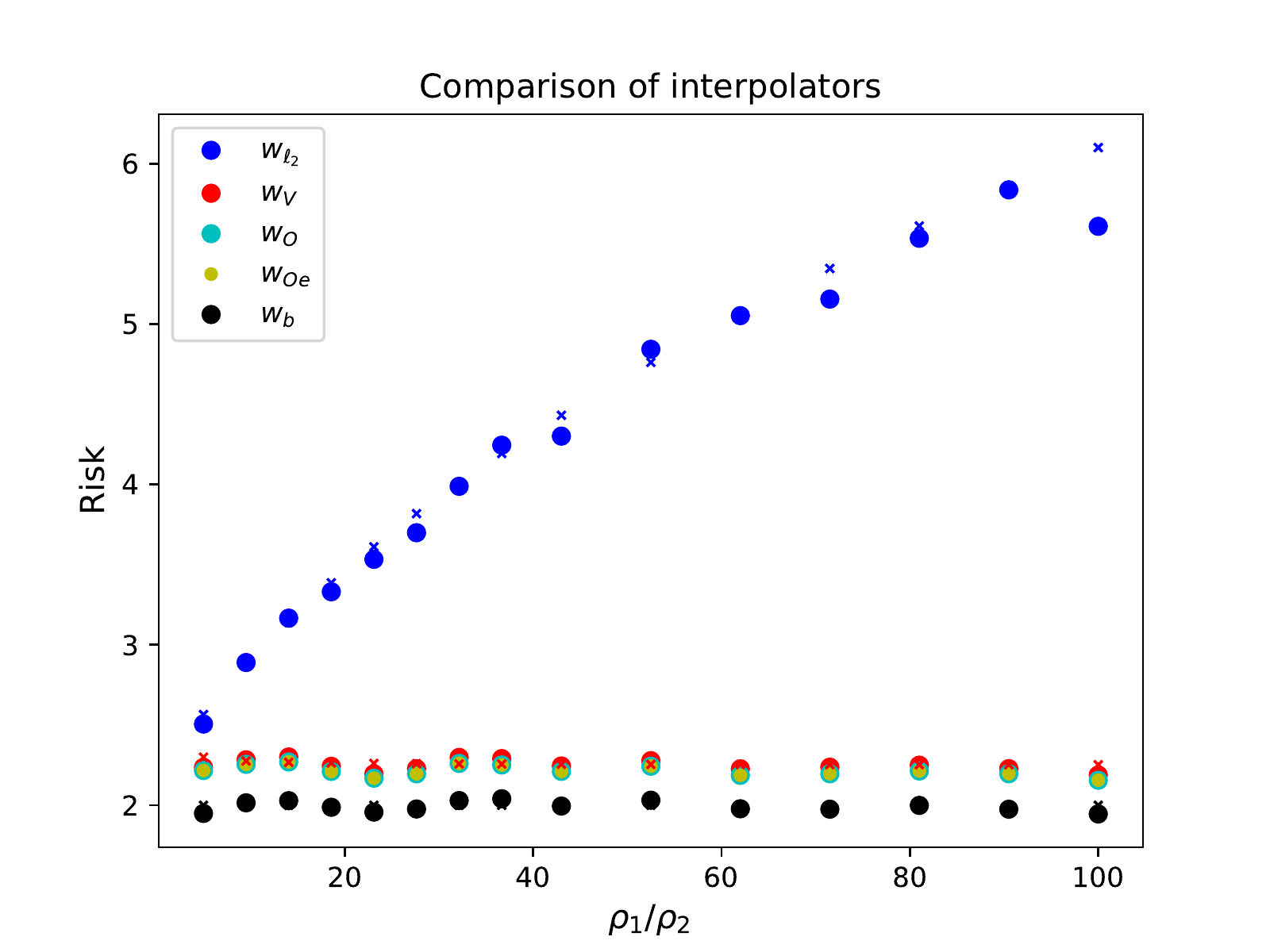} 
        \caption{Plot of $\E_{\xi}r(w)$ (points) for $w\in\{\Wl,\Wc, w_{O}, w_{Oe}, w_b\}$ along with predictions (crosses) from  (\ref{eqn_mirror_asymptotic_risk_bayesian_2}) and (\ref{eqn_asymptotic_risk_general_case}) in the strong weak features model with $r^2 = 1, \sigma^2 = 1, \gamma = 2, \psi_1 = 1/2, n = 3000$ and $\rho_1 = 1$, $\rho_2\to 0$.} \label{fig_variance_to_infty}
    \end{minipage}\hspace{0.5em}
    \begin{minipage}{0.49\textwidth}
        \centering
        \includegraphics[scale=0.45]{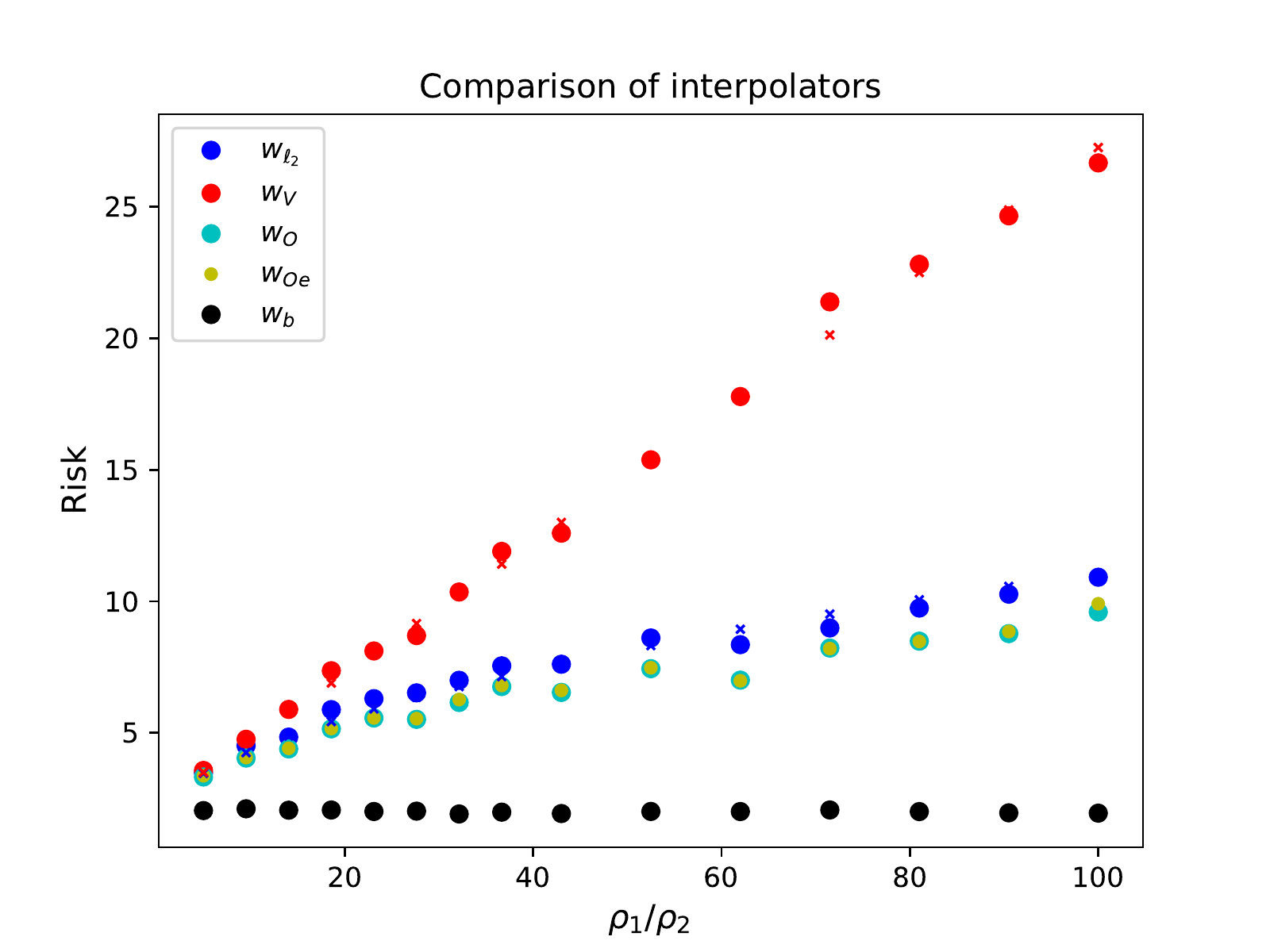} 
        \caption{Plot of $\E_{\xi}r(w)$ (points) for $w\in\{\Wl,\Wc, w_{O}, w_{Oe}, w_b\}$ along with predictions (crosses) from (\ref{eqn_mirror_asymptotic_risk_bayesian_2}) and (\ref{eqn_asymptotic_risk_general_case})  in the strong weak features model with $r^2 = 1, \sigma^2 = 1, \gamma = 2, \psi_1 = 1/2, n = 3000$ and $\rho_2 = 1$, $\rho_1\to\infty$.} \label{fig_bias_to_infty}
    \end{minipage}
\end{figure}
\vspace{-0.1cm}

We note that the empirical estimator $w_{Oe}$ (yellow points), which is a function of only the training data $X$ and $y$ and does not use the population covariance $\Sigma$ or the signal-to-noise ratio $\delta$, performs almost identically to the optimal response-linear achievable interpolator $w_{O}$ (cyan points).\\ 

In this example, we chose $\gamma = 2$ and $
\psi_1 = 1/2$ deliberately. One does not achieve diverging variance for an arbitrary choice of $\gamma$ and $\psi_1$. However, for any $\gamma >1$ such that $\gamma\psi_1 = 1$, the phenomenon of Figure \ref{fig_variance_to_infty} holds (see \ref{appendix_gammapsi_1} of the supplementary material). 
\end{subsection}
\begin{subsection}{\bf Diverging bias of interpolator with optimal variance}\label{section_bias_to_infty}
Now, we illustrate a regime where the best variance interpolator $\Wc$ generalizes arbitrarily worse than $w_{O}$. In the same strong and weak features covariance model described above in Section \ref{subsectionn_variance_gd_to_infty}, when $\gamma = 2$ and $\psi_1 = 1/2$, if we instead have $\rho_1 \to \infty$ and $\rho_2 = 1$, then the asymptotic risk (\ref{auxi_hh}) diverges to infinity linearly. However, the variance of the minimum-norm interpolator in (\ref{variance_gd_strong_weak}) diverges only like $\sqrt{\rho_1}$. Moreover, the bias term satisfies
\begin{align*}
    \Bl &= \frac{r^2}{\gamma v(0)} = \frac{r^2}{\gamma}\sqrt{\rho_1\rho_2}, 
\end{align*}
which also diverges like $\sqrt{\rho_1}$. Now, because $\E_{\xi,w^{\star}} r(w_{O})\leq \E_{\xi,w^{\star}} r(\Wl),$
we have that
\[\lim_{d\to\infty}\E_{\xi,w^{\star}} r(w_{O}) \leq \frac{r^2}{\gamma}\sqrt{\rho_1\rho_2} + \frac{\sigma^2}{2}\bigg(\sqrt{\frac{\rho_1}{\rho_2}} + \sqrt{\frac{\rho_2}{\rho_1}} +3\bigg),\]
so that the asymptotic risk of $w_O$ diverges to infinity as $\sqrt{\rho_1}$. We illustrate this in Figure \ref{fig_bias_to_infty}.\\

We notice that the empirical approximation $w_{Oe}$ again performs in a nearly identical way to the optimal response-linear achievable interpolator $w_{O}$. Moreover, importantly, we note that $\Wc$ and $w_{O}$ are limits of the same algorithm, $w_{t+1} = w_{t} - \eta_t{\bfc}^{-1}\nabla R(w_{t})$, only with different initialization. Hence, this shows that different initialization of the same optimization algorithm can have an arbitrarily large influence on generalization through implicit bias.
\end{subsection}
\end{section}

\section{Random features regression}
The concept of optimal interpolation as a function which is linear in the response variable, is general and can be extended beyond linear models. We present an extension of Proposition \ref{prop_bla} to the setting of random features regression. Random features models were introduced as a random approximation to kernel methods \citep{randomfeatures_rahimi_recht} and can be viewed as a two-layer neural network with first layer randomly initialized and fixed as far as training is concerned. They can be shown to approximate neural networks in certain regimes of training and initialization and hence are often considered in the literature as a first step to address neural networks (e.g.\ \citep{ntk_convergence}). We consider data generated in the same way as before, $y_i = \langle x_i, w^{\star} \rangle + \xi_i$, and the model to be a two-layer neural network $f_a:\Rd\ni x \mapsto a^{T}\sigma(\Theta x/\sqrt{d})$, where the first layer $\Theta\in\R^{N\times d}$ is randomly initialized. This setting, along with $x_i$ and rows of $\Theta$ belonging to the sphere $\mathbb{S}^{d-1}(\sqrt{d})$ with radius $\sqrt{d}$ in $\Rd$, is often considered in the literature on interpolation of random features models \citep{mei2019generalization, montanari_linearized}. If we analogously define the optimal response-linear achievable interpolator in random features regression by 
\begin{align}
  a_{O} = \argmin_{a\in\G\cap\Lc} \E_{\xi,w^{\star}} r(f_a) - r(w^{\star}), \label{eqn_def_bla_rf}
\end{align}
where here $\G = \{a\in\R^{N} : \bfZ a = y\}$ is the set of interpolators, $\bfZ =\sigma(\bfX\Theta^T/\sqrt{d})$ and $\Lc$ is the same as in Definition \ref{def_response-linear_achievable_estimators}, then the following analogue of Proposition \ref{prop_bla} holds.
\begin{proposition}\label{prop_bla_rf}
The optimal response-linear achievable interpolator (\ref{eqn_def_bla_rf}) in random features regression satisfies
\begin{align*}
a_{O} \!=\! \Sigma_z^{-1}\bigg(\Sigma_{zx}\Phi\bfX^{T}
\!+\! \bfZ^{T}\big(\bfZ\Sigma_z^{-1}\bfZ^{T}\big)^{-1}\big(\frac{d}{\delta}I_n \!+\! \bfX\Phi\bfX^{T} \!-\! \bfZ\Sigma_z^{-1}\Sigma_{zx}\Phi\bfX^{T}\big)\bigg)\bigg(\frac{d}{\delta}I_n \!+\! \bfX\Phi\bfX^T\bigg)^{-1}\hspace{-0.2cm}\!y,
\end{align*}
\end{proposition}
Here $\Sigma_z = \E_{\tilde{x}}(\sigma(\Theta \tilde{x}/\sqrt{d})\sigma(\Theta \tilde{x}/\sqrt{d})^T)$ and $\Sigma_{zx} = \E_{\tilde{x}}(\sigma(\Theta \tilde{x}/\sqrt{d})\tilde{x}^T)$ are covariance and cross-covariance matrices, respectively. This interpolator can be again obtained as the implicit bias of preconditioned gradient descent using results of \cite{optimization_geometry}.
\begin{proposition}\label{prop_bla_rf_implicit_bias_of_preconditioned_GD}
The optimal response-linear achievable interpolator (\ref{eqn_def_bla_rf}) in random features regression is the limit of preconditioned gradient descent on the last layer,
\begin{align*}
w_{t+1} = w_{t} - \eta_t{\Sigma_{z}}^{-1}\nabla R(w_{t}),
\end{align*}
provided that the algorithm converges, initialized at
\begin{align*}
a_0 = \Sigma_{z}^{-1}\Sigma_{zx}\Phi\bfX^{T}\bigg(\frac{d}{\delta}I_n + \bfX\Phi\bfX^{T}\bigg)^{-1}y.
\end{align*}
\end{proposition}
In Section \ref{appendix_section_rf_example}, we illustrate the test error of $f_{a}$, with $a = a_{O}$ in comparison to the test error for the minimum-norm interpolator $a = a_{\ell_2} = \bfZ\dg y$ on a standard example.

\section{Conclusion}
In this paper, we investigated how to design interpolators in linear regression which have optimal generalization performance. We designed an interpolator which has optimal risk among interpolators that are a function of the training data, population covariance, signal-to-noise ratio and prior covariance, but does not depend on the true parameter or the noise, where this function is linear in the response variable. We showed that this interpolator is the implicit bias of a covariance-based preconditioned gradient descent algorithm. We identified regimes where other interpolators of interest are arbitrarily worse using computations of their asymptotic risk as $\frac{d}{n} \to \gamma > 1$ with $d,n\to\infty$. \\

In particular, we found a regime where the variance term of the minimum-norm interpolator is arbitrarily large compared to our interpolator. This confirms the phenomenon that implicit bias has an important influence on generalization through the choice of optimization algorithm. \\

We identified a second regime where the interpolator that has best variance is arbitrarily worse than our interpolator. In this second example, both interpolators are the implicit bias of the same algorithm, but with different initialization. This contributes to illustrating that initialization has an important influence on generalization. \\

We also considered an empirical approximation of the optimal response-linear achievable interpolator, which uses only the training data $X$ and $y$ and does not assume knowledge of the population covariance matrix, the signal-to-noise ratio or the prior covariance and empirically observe that it generalizes in a nearly identical way to the optimal response-linear achievable interpolator in the examples that we consider.\\

A limitation of this work includes a precise guarantee on the approximation error of the Graphical Lasso for a general covariance matrix $\bfc$. Some guarantees are in \citep{ravikumar2008highdimensional}, however establishing guarantees for a general covariance matrix would be a contribution on its own. \\

A natural question for future research, which also motivated our work, is how to systematically design new ways of interpolation, which are adapted to the distribution of the data and related to notions of optimality, for more general and complex overparametrized machine learning models such as neural networks.

\section{Acknowledgements}
The authors would like to thank Dominic Richards, Edgar Dobriban and the anonymous reviewers for valuable insights which contributed to the technical quality of the paper. Patrick Rebeschini was supported in part by the Alan Turing Institute under the EPSRC grant EP/N510129/1.

\bibliography{bibli}

\newpage

\appendix
\section{Supplementary Material}
\begin{subsection}{Proof of Propostion \ref{prop_bla}}\label{appendix_proof_bla}
We prove that the optimal response-linear achievable interpolator in linear regression is
\[w_{O} = \bigg(\frac{\delta}{d}\bfm\bfX^T \!+\! \cmhalf(\bfX\cmhalf)\dg\bigg)\bigg( I_n \!+\! \frac{\delta}{d}\bfX\bfm\bfX^T\bigg)^{-1}\!\!y.\]
\begin{proof}
First, we note that as $w_{O}\in\Lc$, there exists $\R^{d\times n} \ni \bfq = \bfq(\bfX, \bfc, \bfm, \delta)$ such that $w_{O} = \bfq y = \bfq \bfX w^{\star} + \bfq \xi$. Therefore, the definition of $w_{O}$,
\begin{align*}
  w_{O} = \argmin_{w\in\G\cap\Lc} \E_{\xi,w^{\star}} r(w) - r(w^{\star}),
\end{align*}
can be restated as
\begin{align}
w_{O} = \argmin_{\substack{w_{O} = \bfq y \\ \bfX\bfq y = y}}B(w_{O}) + V(w_{O}), \label{eqn_first_restatement_main_prop}
\end{align}
where
\begin{align*}
B(w_{O}) &= \frac{r^2}{d}\text{Tr}\big(\bfc(\bfq\bfX - I_d)\bfm(\bfq\bfX - I_d)^T\big),\\
V(w_{O}) &= \sigma^2\text{Tr}\big(\bfc\bfq\bfq^T\big).
\end{align*}

\begin{claim}\label{claim_in_proof_of_prop_1}
(\ref{eqn_first_restatement_main_prop}) implies 
\begin{align}
w_{O} = \argmin_{\substack{w_{O} = \bfq y \\ \bfX\bfq \bfX = \bfX}}B(w_{O}) + V(w_{O}). \label{eqn_second_restatement_main_prop}
\end{align}
\end{claim}
We prove this claim. $\bfX\bfq y = y$ almost surely for all realizations of the data (that is, a.s. for all realizations of $\bfX, \xi, w^{\star}$) implies
\begin{align*}
    0 = \E\big( \bfX\bfq(\bfX w^{\star} + \xi) | \bfX, w^{\star} \big) - \E\big( \bfX w^{\star} + \xi | \bfX, w^{\star} \big) = \bfX(\bfq\bfX - I_d)w^{\star}
\end{align*}
almost surely for all realizations of $\bfX, w^{\star}$. Therefore,
\begin{align}
    0 = \E_{w^{\star}}\big( \Vert \bfX(\bfq\bfX - I_d)w^{\star} \Vert_2^2 \;|\; w^{\star}\big) = {w^{\star}}^T\E\big((\bfq\bfX - I_d)^T\bfX^T\bfX(\bfq\bfX - I_d)\big)w^{\star} \label{appendix_auxininus_equationinus}
\end{align}
almost surely for all realizations of $w^{\star}$. It follows that $\E\big((\bfq\bfX - I_d)^T\bfX^T\bfX(\bfq\bfX - I_d)\big) = 0$. This is because, if not, then there exists $v\in\Rd$ and $\epsilon >0$ such that 
\begin{equation}
\forall u\in B_{\epsilon}(v) \qquad u^T\E\big((\bfq\bfX - I_d)^T\bfX^T\bfX(\bfq\bfX - I_d)\big)u > 0. \label{appendix_auxius_equationius}
\end{equation}
Recall that $w^{\star}\sim {\mathcal{P}_{w^{\star}}}$, where ${\mathcal{P}_{w^{\star}}}$ is by assumption such that $\nu(A) > 0$ implies ${\mathcal{P}_{w^{\star}}}(A) > 0 $ for all Lebesgue measurable $A\in\Rd$ (where $\nu$ is the Lebesgue measure). However, as $B_{\epsilon}(v)$ has positive Lebesgue measure, (\ref{appendix_auxius_equationius}) is hence a contradiction to (\ref{appendix_auxininus_equationinus}). Finally, as $\Vert \bfa \Vert = \sqrt{\text{Tr}(\bfa\bfa^T)}$ is a norm, $\E\big((\bfq\bfX - I_d)^T\bfX^T\bfX(\bfq\bfX - I_d)\big) = 0$ implies that $\bfX(\bfq\bfX - I_d) = 0$ almost surely. This proves the claim.\\

Now we use Theorem 2 of \cite{penrose_pseudoinv_def} which states that for any matrices $\bfa, \bfb, \bfcc$ and $\bfd$, all solutions $\bfb$ to the equation $\bfa \bfb \bfcc = \bfd$ can be written as $\bfb = \bfa\dg\bfd\bfcc\dg + \bfz - \bfa\dg\bfa\bfz\bfcc\bfcc\dg$ where $\bfz$ is arbitrary. Therefore, $\bfX\bfq \bfX = \bfX$ is equivalent to
\begin{align*}
\bfq = \bfX\dg\bfX\bfX\dg + \bfz - \bfX\dg\bfX\bfz\bfX\bfX\dg = \bfX\dg + \bfz - \bfX\dg\bfX\bfz
\end{align*} 
for some arbitrary $\bfz\in\R^{d\times n}$. Hence, if we write $Q = Q(S)$ and $w_{O} = w_{O}(S)$, (\ref{eqn_second_restatement_main_prop}) is equivalent to an unconstrained optimization problem over $\R^{d\times n}$ in the form
\begin{align}
w_{O} = \argmin_{\bfz}f(\bfz), \label{eqn_third_restatement_main_prop}
\end{align}
where $f(\bfz) = B(w_{O}(\bfz)) + V(w_{O}(\bfz))$. Now we show that $f:\R^{d\times n} \to \R$ is strictly convex. Note that the map
\[\R^{d\times n}\ni\bfz\mapsto \chalf(\bfq(\bfz)\bfX - I_d)\mhalf\]
is affine and nonzero and the map $\Rdd\ni\bfa\mapsto\text{Tr}(\bfa\bfa^T)$ is strictly convex because $\Vert \bfa \Vert = \sqrt{\text{Tr}(\bfa\bfa^T)}$ is a norm. The composition of these two maps is $\R^{d\times n}\ni \bfz\mapsto \frac{d}{r^2}B(w_{O}(\bfz))$, which is therefore strictly convex. A similar argument proves that $\R^{d\times n}\ni \bfz\mapsto V(w_{O}(\bfz))$ is strictly convex and hence also $f$ is. Moreover, $f$ is differentiable. Therefore, to find a unique global minimum of $f$, it is enough to find $\bfz^{\star}\in\R^{d\times n}$ such that $\partial f(\bfz^{\star}) = 0$. Using tools of matrix calculus we find
\begin{align*}
\partial f(\bfz) = 2(I_d - \bfX\dg\bfX)\bfc A,
\end{align*}
where
\begin{align*}
A &= \bigg(\sigma^2\bfz +\frac{r^2}{d}(\bfz\bfX - I_d)\bfm\bfX^T+\bfX\dg(I_n - \bfX\bfz)(\sigma^2I_n +\frac{r^2}{d}\bfX\bfm\bfX^T)\bigg).    
\end{align*}
Because $\Rd \ni v \mapsto (I_d - \bfX\dg\bfX)v$ is the projection onto $\text{Ker}(\bfX) = \text{Im}(\bfX^T)^{\bot}$, this hints towards finding $\bfz^{\star}$ such that $A = \bfc^{-1}\bfX^TB$ for some matrix $B$. This is achieved, for example, if
\begin{align*}
\sigma^2\bfz^{\star} +\frac{r^2}{d}(\bfz^{\star}\bfX - I_d)\bfm\bfX^T = \bfc^{-1}\bfX^TB
\end{align*}
and
\begin{align*}
I_n - \bfX\bfz^{\star} = 0,
\end{align*}
for some matrix $B$. Putting the two equations together implies $B = \sigma^2(\bfX\bfc^{-1}\bfX^T)^{-1}$ and hence, using that $\cmhalf(\bfX\cmhalf)\dg = \bfc^{-1}\bfX^{T}(\bfX\bfc^{-1}\bfX^T)^{-1}$ and $\delta = \frac{r^2}{\sigma^2}$, we have
\[\bfz^{\star} = \bigg(\frac{\delta}{d}\bfm\bfX^T + \cmhalf(\bfX\cmhalf)\dg\bigg)\bigg( I_n + \frac{\delta}{d}\bfX\bfm\bfX^T\bigg)^{-1}.\]
Finaly, because $\bfX\bfz^{\star} = I_n$, it follows that $\bfq^{\star} = \bfX\dg + \bfz^{\star} - \bfX\dg\bfX\bfz^{\star} = \bfz^{\star}$ and hence
\begin{align*}
w_{O} = \bfq^{\star}y = \bigg(\frac{\delta}{d}\bfm\bfX^T \!+\! \cmhalf(\bfX\cmhalf)\dg\bigg)\bigg( I_n \!+\! \frac{\delta}{d}\bfX\bfm\bfX^T\bigg)^{-1}\!\!y.
\end{align*}
\end{proof}
\end{subsection}

\begin{subsection}{Proof of Proposition \ref{prop_bla_implicit_bias_of_preconditioned_GD}}\label{appendix_subsection_proof_prop_bla_implicit_bias}
We prove that the optimal response-linear achievable interpolator $w_{O}$ is the limit of preconditioned gradient descent
\begin{equation}
w_{t+1} = w_{t} - \eta_t{\bfc}^{-1}\nabla R(w_{t}), \label{eqn_preconditioned_gd_update_rule_appendix}
\end{equation}
provided that the algorithm converges, initialized at
\begin{align*}
w_0 = \frac{\delta}{d}\bfm\bfX^{T}\bigg( I_n + \frac{\delta}{d}\bfX\bfm\bfX^T\bigg)^{-1}\! y.
\end{align*}

\begin{proof}
Preconditioned gradient descent (\ref{eqn_preconditioned_gd_update_rule_appendix}) is equivalent to mirror descent 
\begin{align*}
\nabla \phi(w_{t+1}) = \nabla \phi(w_{t}) - \eta_t\nabla R(w_{t}) 
\end{align*}
with mirror map $\phi(w) = \frac{1}{2}w^T \bfc w$. By a result of \citep{optimization_geometry}, if mirror descent with mirror map $\phi$, a unique root loss function (e.g. the squared error loss), initialisation $w_0$ and stepsize $(\eta_t)_{t\in \mathbb{N} }$ satisfies $\lim_{t\to\infty}R(w_{t}) = 0$ then
 \begin{align*}
 \lim_{t\to\infty}w_{t} = \argmin_{w\in \G} D_{\phi}(w,w_0),\label{no constraint minim problem}
 \end{align*} 
 where 
 \[D_{\phi}(w,w_0) = \phi(w) - \phi(w_0) - \nabla\phi(w_0)^T(w-w_0)\]
 is the associated Bregman divergence. By this result applied with $\phi(w) = \frac{1}{2}w^T \bfc w$, we have that if preconditioned gradient descent (\ref{eqn_preconditioned_gd_update_rule_appendix}) initialized at $w_0$ converges, its limit satisfies
\begin{align*}
\lim_{t\to\infty}w_{t} & = \argmin_{w\in\Rd\,:\,\bfX w = y} \Vert \chalf (w - w_0) \Vert_2^2. \nonumber
\end{align*}
After a linear transformation and an application of a result about approximate solutions to linear matrix equations \citep{penrose_lin_eqns}, similarly as in (\ref{explicit_form_of_best_parameter}), we obtain
\begin{align}
\lim_{t\to\infty}w_{t} = \cmhalf (\bfX\cmhalf)\dg (y - \bfX w_0) + w_0.  \label{eqn_impl_bias_md_general}
\end{align}
Finally, using
\begin{align*}
w_0 = \frac{\delta}{d}\bfm\bfX^{T}\bigg( I_n + \frac{\delta}{d}\bfX\bfm\bfX^T\bigg)^{-1}\! y,
\end{align*}
we obtain
\begin{align*}
\cmhalf (\bfX\cmhalf)\dg (y - \bfX w_0) + w_0 = \bigg(\frac{\delta}{d}\bfm\bfX^T \!+\! \cmhalf(\bfX\cmhalf)\dg\bigg)\bigg( I_n \!+\! \frac{\delta}{d}\bfX\bfm\bfX^T\bigg)^{-1}\!\!y = w_{O}. 
\end{align*}
\end{proof}
\end{subsection}

\begin{subsection}{Proof of Proposition \ref{prop_preconditioned_GD_achieves_optimal_variance}}\label{supplementary_material_proof_Proposition_1}
We prove that for any deterministic initialization $w_0\in\Rd$, the limit of converging preconditioned gradient descent $w_{t+1} = w_{t} - \eta_t{\bfc}^{-1}\nabla R(w_{t})$
satisfies that
\begin{align*}
\lim_{t\to\infty}w_{t} = \argmin_{w\in\G} V(w). 
\end{align*}
\begin{proof}
Recall from (\ref{eqn_impl_bias_md_general}) that
\begin{align*}
\lim_{t\to\infty}w_{t} = \cmhalf (\bfX\cmhalf)\dg (y - \bfX w_0) + w_0,
\end{align*}
and the definition of the variance $V(w) = \E_{\xi, w^{\star}}\Vert w - \E(w|w^{\star}, X) \Vert^2_{\bfc}$. Therefore, we have
\[\lim_{t\to\infty}w_{t} - \E(\lim_{t\to\infty}w_{t}|w^{\star}, \bfX) = \cmhalf (\bfX\cmhalf)\dg\xi. \]
Moreover, the optimal interpolator of Definition \ref{best_possible_interpolator} satisfies
\[w_b = w^{\star} + \cmhalf (\bfX\cmhalf)\dg\xi,\]
so that
\[w_b - \E(w_b|w^{\star}, \bfX) = \cmhalf (\bfX\cmhalf)\dg\xi\]
and hence
\[V(w_b) = V(\lim_{t\to\infty}w_{t}) = \E_{\xi}\Vert \cmhalf (\bfX\cmhalf)\dg\xi \Vert^2_{\bfc}.\]
In other words, $\lim_{t\to\infty}w_{t}$ fits the noise in exactly the same way as the optimal interpolator $w_b$, which has the smallest possible risk among all interpolators. Hence, it is enough to show that $w_b$ also has smallest possible variance among all interpolators. We argue by contradiction. Assume that $\Wh$ is an interpolator with smaller variance than $w_b$. Then
\[X\Wh = y\]
implies that
\[X\big(\Wh - \E(\Wh | w^{\star}, \bfX)\big) = \xi\]
and hence $w^{\star} + \Wh - \E(\Wh | w^{\star}, \bfX)$ is also an interpolator. But $w^{\star} + \Wh - \E(\Wh | w^{\star}, \bfX)$ has zero bias (recall that the definition of bias is $B(w) = \E_{\xi, w^{\star}}\Vert \E(w|w^{\star}, \bfX) - w^{\star} \Vert^2_{\bfc}$) and therefore, by assumption, has smaller risk than $w_b$. This is a contradiction. 
\end{proof}
\end{subsection}

\begin{subsection}{Assumption \ref{assumption_distr_x} implies $\text{rank}(X) = n$ with probability $1$.}\label{appendix_ass_1_implies_rank_1}
Note that
\begin{align*}
    \Pp(\text{rank}(X) \ne n) &= \Pp(|\text{Span}(x_1, \dots, x_n)| < n )\\
    &= \Pp(\cup_{i\in\{1, \dots,n\}}\{x_i \in \text{Span}(x_1, \dots, x_{i-1}, x_{i+1}, \dots, x_n)\})\\
    &\leq \sum_{i=1}^n\Pp(\{x_i \in \text{Span}(x_1, \dots, x_{i-1}, x_{i+1}, \dots, x_n)\})\\
    &= \sum_{i=1}^n\E\big(\Pp(\{x_i \in \text{Span}(x_1, \dots, x_{i-1}, x_{i+1}, \dots, x_n)\}| x_1, \dots, x_{i-1}, x_{i+1}, \dots, x_n)\big)\\
    &= 0,
\end{align*}
where the last equation follows directly by applying Assumption \ref{assumption_distr_x}.
\end{subsection}

\begin{subsection}{Response-linear interpolator with optimal bias}\label{appendix_interpolator_with_optimal_bias}
By choosing $\sigma^2 = 0$ in \ref{appendix_proof_bla}, the proof of Proposition \ref{prop_bla}, one obtains the interpolator with optimal bias among response-linear achievable interpolators. This interpolator is
\[\Phi\bfX^T(\bfX\Phi\bfX^T)^{-1}y,\]
which is in agreement with the asymptotic result of \citep{amari2020does}. Therefore, when the prior is isotropic, as claimed in \ref{section_diverging_bias} the interpolator with optimal bias among response-linear achievable interpolators is the minimum-norm interpolator.
\end{subsection}

\begin{subsection}{Proof of equation (\ref{eqn_mirror_asymptotic_risk_bayesian_2})}
We prove that, when the prior is isotropic $\Phi = I_d$, under Assumptions \ref{ass_upper_boudn_on_eval} and \ref{ass_convergence_spectral_ditrib}, if $n,d\to\infty$ with $\frac{d}{n}\to\gamma > 1$ then we have with probability $1$ that
\begin{align*}
\lim_{d\to\infty}\E_{\xi, w^{*}} r(\Wc) - r(w^{\star}) = B(\Wc) + V(\Wc), 
\end{align*}
where
\begin{align*}
    B(\Wc) &= r^2\frac{\gamma-1}{\gamma}\int_{0}^{\infty}s \; d\mathcal{H}(s),\\
    V(\Wc) &= \frac{\sigma^2}{\gamma - 1}.
\end{align*}
\begin{proof}
The proof uses techniques which were already developed in \cite{hastie2019surprises_double_descent}. Namely Theorem 1 of \cite{rubio_mestre} and an exchange of limits. Recall that
\begin{align*}
\Wc &= \cmhalf (\bfX\cmhalf)\dg \bfX w^{\star} + \cmhalf (\bfX\cmhalf)\dg\xi 
\end{align*}
and $\bfZ = \bfX\bfc^{-\frac{1}{2}}$. Therefore, we have
\begin{align*}
\E_{\xi,w^{\star}}r(\Wc) - r(w^{\star}) = B(\Wc) + V(\Wc),
\end{align*}
where, it was proved in \cite{hastie2019surprises_double_descent} that
\begin{align*}
V(\Wc) = \sigma^2\text{Tr}\big((\bfZ^T \bfZ)^{\dagger}\big) \longrightarrow \frac{\sigma^2}{\gamma - 1}
\end{align*}
because $\bfZ_i \overset{\text{i.i.d}}{\sim}\mathcal{N}(0,I_d)$. For the bias term we also use techniques similar to \citep{hastie2019surprises_double_descent}. In particular, we have
\begin{align*}
B(\Wc) &= \E_{w^{\star}} {w^{\star}}^T \chalf (I - \bfZ\dg\bfZ)^T (I - \bfZ\dg\bfZ){\chalf w^{\star}} \\
&= \E_{w^{\star}} \text{Tr}\big(\chalf w^{\star}{w^{\star}}^T\chalf(I - \bfZ\dg\bfZ)\big)\\
&= \frac{r^2}{d} \text{Tr}\big(\bfc(I - \bfZ\dg\bfZ)\big).
\end{align*}
Moreover, we have
\begin{align*}
\bfZ\dg &= (\bfZ^T\bfZ)\dg\bfZ^T,
\end{align*}
so that if we denote $\widehat{\Sigma} = \bfZ^T\bfZ/n$ to be the empirical covariance matrix of the whitened features, then
\begin{align*}
\bfZ\dg\bfZ & = \widehat{\Sigma}\dg\widehat{\Sigma} = \lim_{\lambda \to 0^{+}}(\widehat{\Sigma} + \lambda I_d)^{-1}\widehat{\Sigma}.
\end{align*}
Therefore, 
\begin{align*}
B(\Wc) &= \lim_{\lambda \to 0^{+}}\frac{r^2}{d} \text{Tr}\big((I - (\widehat{\Sigma} + \lambda I_d)^{-1}\widehat{\Sigma} )\bfc\big)\\
&= \lim_{\lambda \to 0^{+}}\frac{r^2}{d} \lambda\text{Tr}\big( (\widehat{\Sigma} + \lambda I_d)^{-1}\bfc\big).
\end{align*}
Now, we use Theorem 1 of \citep{rubio_mestre} to compute the limit of $\text{Tr}\big( (\widehat{\Sigma} + \lambda I_d)^{-1}\bfc\big)$ as $d/n\to\gamma>1$ with $d\to\infty, n\to\infty$. 
This theorem shows that if $\Theta = (\Theta_d)_{d\in\N}$ is a sequence of matrices such that $\sqrt{\text{Tr}\big(\Theta\Theta^T\big)}$ is uniformly bounded, then
\begin{align}
\text{Tr}\big(\Theta((\widehat{\Sigma} + \lambda I_d)^{-1} - c_d(\lambda) I_d)\big) \longrightarrow 0, \label{rubio_mestre_convergence_theta}
\end{align}
where $c_d(\lambda)$ is a certain quantity defined through an implicit equation (for simplicity we do not define it, as we only need to know its limit). If we choose $\Theta = I_d/d$, then because $\widehat{\Sigma} = \bfZ^T\bfZ/n$ where $Z_i\overset{\text{i.i.d}}{\sim}\mathcal{N}(0,I_d)$ and the spectral distribution $\mathcal{F}_{I_d}$ is just the distribution induced by the measure $\delta_1$ for all $d\in\N$, we have 
\[\lim_{d\to\infty}c_d(\lambda) \to m(-\lambda),\]
where $m$ is the Stieltjes transform of the limiting spectral distribution of $\bfZ^T\bfZ/n$ given by the Mar{\v c}enko-Pastur theorem \citep{marchenko-pastur}. Now that we know $c_d(\lambda) \to m(-\lambda)$, we use (\ref{rubio_mestre_convergence_theta}) again but with $\Theta = \bfc/d$. This shows that
\begin{align*}
\text{Tr}\bigg(\frac{\bfc}{d}(\widehat{\Sigma} + \lambda I_d)^{-1}\bigg) - \text{Tr}\bigg(\frac{\bfc}{d}\bigg)m(-\lambda) \longrightarrow 0,
\end{align*}
provided that $\sqrt{\text{Tr}(\bfc^2)}/d$ is uniformly bounded. This is true when Assumption \ref{ass_upper_boudn_on_eval} holds so that $\lambda_{\text{max}}(\bfc)$ is uniformly bounded. Moreover, 
\begin{align*}
\frac{1}{d}\text{Tr}(\bfc) &= \frac{1}{d}\sum_{i=1}^{d}\lambda_i(\bfc) = \int s \;d\mathcal{F}_{\bfc}(s)
\longrightarrow \int s\; d\mathcal{H}(s),
\end{align*}
where in the last line we used Assumptions \ref{ass_convergence_spectral_ditrib} and \ref{ass_upper_boudn_on_eval}. Therefore, we arrive at 
\begin{align*}
\frac{r^2}{d}\lambda\text{Tr}\big(\bfc(\widehat{\Sigma} + \lambda I_d)^{-1}\big) \longrightarrow r^2\lambda m(-\lambda)\int s\; d\mathcal{H}(s).
\end{align*}
Finally, assuming we can exchange limits (which we justify shortly), we have
\begin{align}
    \lim_{d\to\infty}B(\Wc) &= \lim_{d\to\infty}\lim_{\lambda\to 0^{+}} \frac{r^2}{d}\lambda\text{Tr}\big(\bfc(\widehat{\Sigma} + \lambda I_d)^{-1}\big) \label{exchanged_limits_auxi}\\
    &= \lim_{\lambda\to 0^{+}}\lim_{d\to\infty} \frac{r^2}{d}\lambda\text{Tr}\big(\bfc(\widehat{\Sigma} + \lambda I_d)^{-1}\big) \nonumber \\
    &= \lim_{\lambda\to 0^{+}} r^2\lambda m(-\lambda)\int s\; d\mathcal{H}(s), \nonumber
\end{align}
and because $m$ is the Stieltjes transform of the standard Mar{\v c}enko-Pastur law, it is known (Proposition 3.11 of \cite{bai_silverstein}) that
\[\lim_{\lambda\to 0^{+}} \lambda m(-\lambda) = \frac{\gamma - 1}{\gamma}.\]
However, to fully finish the proof, one needs to first justify exchanging the limits in (\ref{exchanged_limits_auxi}). We do this now. Define a sequence of functions $f_d:\R^{+} \to \R$ with 
\begin{align*}
    f_d(\lambda) = \frac{r^2}{d}\lambda\text{Tr}\big(\bfc(\widehat{\Sigma} + \lambda I_d)^{-1}\big)
\end{align*}
and 
\begin{align*}
    f(\lambda) = r^2\lambda m(-\lambda)\int s\; d\mathcal{H}(s).
\end{align*}
We proved that $\lim_{d\to\infty}f_d(\lambda) = f(\lambda)$ pointwise. To assert
\begin{align*}
    \lim_{d\to\infty} \lim_{\lambda \to 0^{+}} f_d(\lambda) = \lim_{\lambda \to 0^{+}} f(\lambda)
\end{align*}
it is therefore, by the Moore-Osgood theorem, enough to show that $(f_d)_{d\in\N}$ is uniformly convergent. As $(f_d)_{d\in\N}$ has a pointwise limit, it is enough to show that every subsequence of $(f_d)_{d\in\N}$ has a uniformly convergent subsequence. For this, we show that $(f_d)_{d\in\N}$ is uniformly bounded and has uniformly bounded derivative, which gives the convergent subsequences by the Arzela-Ascoli theorem. Indeed, we have that 
\begin{align*}
    |f_d(\lambda)| \leq r^2\lambda_{\text{max}}(\bfc)
\end{align*}
and as $f_d'(\lambda) = \frac{r^2}{d}\text{Tr}\big(\bfc(\widehat{\Sigma} + \lambda I_d)^{-2}\widehat{\Sigma}\big)$ we have
\begin{align*}
    |f_d'(\lambda)| &\leq r^2\lambda_{\text{max}}(\bfc)\frac{\lambda_{\text{max}}(\widehat{\Sigma})}{(\lambda_{\text{min}}(\widehat{\Sigma})^{+} + \lambda)^2} \leq r^2\lambda_{\text{max}}(\bfc)8\frac{(\sqrt{\gamma}+1)^2}{(\sqrt{\gamma}-1)^4}.
\end{align*}
In the inequality we used Theorem 1 of \cite{baiyin} which shows that, with probability 1, 
\begin{align*}
\liminf_{d\to\infty}\lambda_{\textrm{min}}(\widehat{\Sigma})^{+} &\geq \frac{1}{2}(\sqrt{\gamma} - 1)^2,\\
\limsup_{d\to\infty}\lambda_{\textrm{max}}(\widehat{\Sigma}) &\leq 2(\sqrt{\gamma} + 1)^2.
\end{align*}
\end{proof}
\end{subsection}

\begin{subsection}{$\Vl$ in strong weak features model with $\gamma\psi_1 = 1$.}\label{appendix_gammapsi_1}
In this subsection we justify the statement (of the last paragraph of Section \ref{subsectionn_variance_gd_to_infty}) that, in the strong weak features model of covariance matrices
\begin{align*}
\bfc = \text{diag}(\rho_1,\dots, \rho_1, \rho_2, \dots, \rho_2)\in\Rdd,  
\end{align*}
where the number of $\rho_1$s is $d\cdot\psi_1$ with $\psi_1 \in [0,1]$, we have
\begin{align*}
\Vl &= \sigma^2\bigg(\frac{v'(0)}{v(0)^2}-1\bigg) \to \infty    
\end{align*}
as $\rho_2 \to 0$ for any $\gamma>1$ such that $\gamma\psi_1 = 1$. Indeed, using the relation 
\begin{align*}
m(z) + \frac{1}{z} = \gamma(v(z) + \frac{1}{z}),    
\end{align*}
(which can be shown to hold) and Definition \ref{definition_Stieltjes_transform} of the Stieltjes transform and its limit, it can be checked that 
\begin{align*}
    v(0) = \frac{x + \sqrt{x^2 +4(\gamma -1)\rho_1\rho_2}}{2(\gamma - 1)\rho_1\rho_2},
\end{align*}
where $x = \rho_1 + \rho_2 - \gamma\psi_1\rho_1 - \gamma(1-\psi_1)\rho_2$. Moreover, taking a derivative in the Silverstein equation \citep{silverstein1995}, which states that
\begin{align*}
    -\frac{1}{v(z)} = z - \gamma\int\frac{s}{1+sv(z)}\;d\mathcal{H}(s),
\end{align*}
gives
\begin{align}
    \frac{v'(0)}{v(0)^2} - 1 = \gamma v'(0)\Delta \label{supplementary_eqn_auxinko},
\end{align}
where 
\begin{align}
\Delta = \bigg(\frac{\psi_1\rho_1^2}{(1+\rho_1v(0))^2} + \frac{(1-\psi_1)\rho_2^2}{(1+\rho_2v(0))^2}\bigg). \label{supplementary_eqn_auxinko_2}
\end{align}
By rearranging (\ref{supplementary_eqn_auxinko}) we obtain
\begin{align}
\frac{v'(0)}{v(0)^2} - 1 = \frac{1}{1-\gamma\Delta v(0)^2} - 1.   \label{supplementary_eqn_auxinko_3}
\end{align}
Now if $\gamma\psi_1 = 1$, then $x = \rho_2(2 - \gamma)$ and
\begin{align*}
   v(0) &= \frac{2-\gamma + \sqrt{(2-\gamma)^2+4(\gamma-1)\frac{\rho_1}{\rho_2}}}{2(\gamma-1)\rho_1}.
\end{align*}
Hence
\begin{align*}
v(0)\sqrt{\rho_2}\longrightarrow\sqrt{\frac{1}{(\gamma-1)\rho_1}}   
\end{align*}
as $\rho_2\to 0$. Using (\ref{supplementary_eqn_auxinko_2}) and (\ref{supplementary_eqn_auxinko_3}), it can be therefore checked that, as $\rho_2\to 0$,
\begin{align*}
    \sqrt{\rho_2}\bigg(\frac{v'(0)}{v(0)^2} - 1\bigg) \longrightarrow \frac{1}{2}\sqrt{\frac{\rho_1}{\gamma-1}}.
\end{align*}
Therefore, $\frac{v'(0)}{v(0)^2} - 1 \to \infty$ as $\rho_2\to 0$.
\end{subsection}

\begin{subsection}{Empirical comparison of the Graphical Lasso for some covariance matrices}\label{appendix_empirical_comparison}
We illustrate how the interpolator $w_{Oe}$, obtained by using the Graphical Lasso approximation of the covariance matrix, performs in comparison to the optimal response-linear achievable interpolator $w_{O}$ for two regimes of covariance matrices. In this Section, we do this in the regime of an isotropic prior $\bfm = I_d$. See Section \ref{subsection_non_isotropic} for the case $\bfm \ne I_d$. The interpolator 
\begin{align*}
w_{Oe} & \!=\! \bigg(\frac{\delta_e}{d}\bfX^T \!+\! \cemhalf(\bfX\cemhalf)\dg\bigg)\bigg( I_n \!+\! \frac{\delta_e}{d}\bfX\bfX^T\bigg)^{-1}\!\!y,
\end{align*}
is constructed by using the Graphical Lasso estimator $\bfc_e$ \citep{glasso} of the covariance matrix (implemented in scikit-learn \citep{scikit-learn}), and choosing $\delta_e$ which minimizes the crossvalidated error on random subsets of the data as described in Section \ref{section_approximation_of_the_true_covariance}.

First, we look at the autoregressive regime, where 
\begin{align*}
    \bfc_{i,j} = \rho^{|i-j|}
\end{align*}
for all $i,j \in \{1,\dots,d\}$ and $\rho\in (0,1)$.

\vspace{-0.4cm}
\begin{figure}[H]
\begin{center}
\includegraphics[scale=0.55]{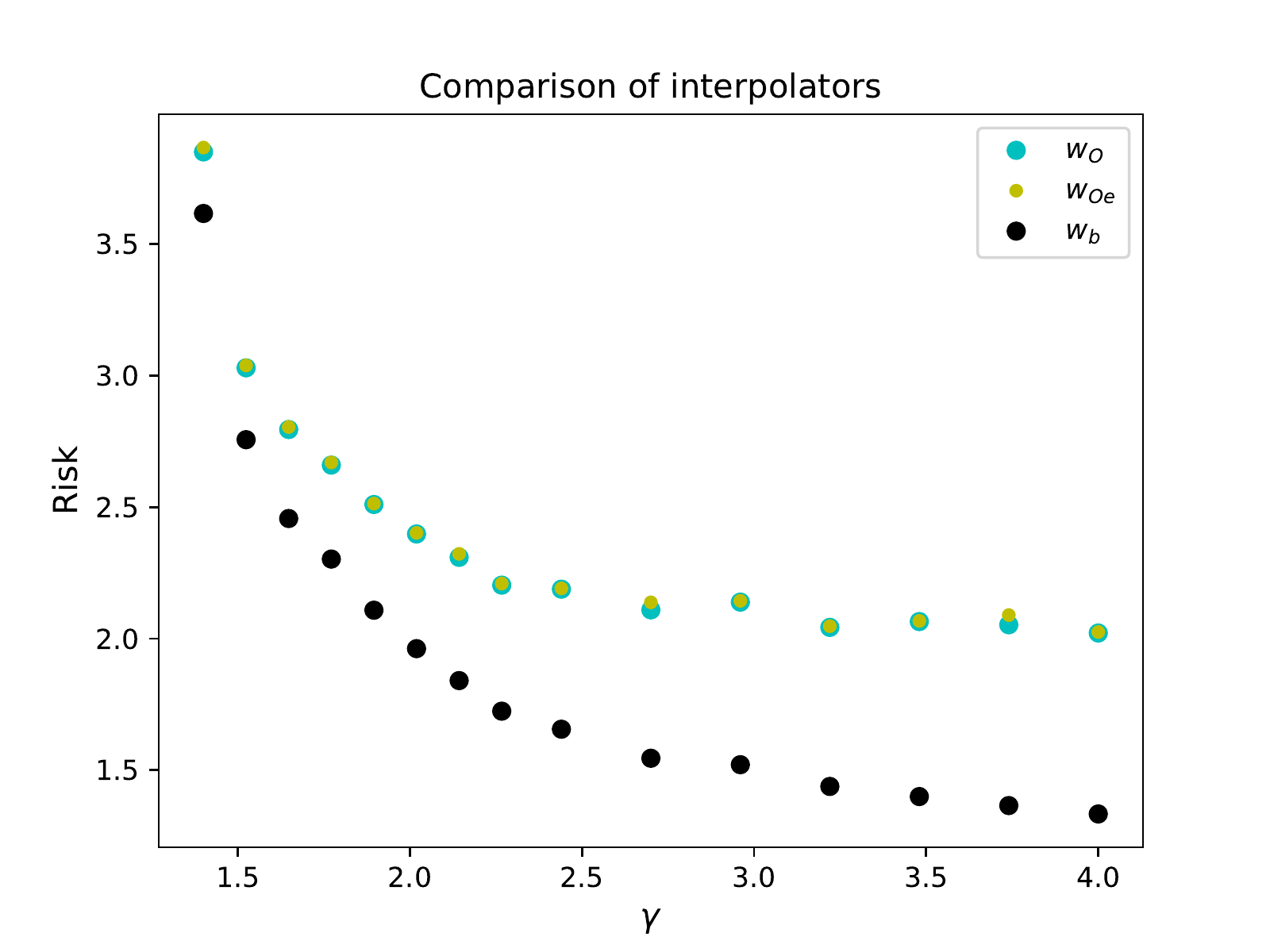}
\caption{Plot of $\E_{\xi}r(w)$ (points) for $w\in\{w_{O}, w_{Oe}, w_b\}$ in the autoregressive regime with $d = \lfloor \gamma n \rfloor, r^2 = 1, \sigma^2 = 1, n = 2000, \rho = 0.5. $} 
\label{fig_autoregressive}
\end{center}
\end{figure}
\vspace{-0.45cm}

Second, we consider an exponential regime \citep{dobriban}, where the eigenvalues of $\bfc$ are evenly spaced quantiles of the standard exponential distribution. Namely,
\begin{align*}
    \bfc_{i,i} = -\text{log}(1-p_i),
\end{align*}
where $p_i = i/(d+1)\in(0,1)$ for $i\in\{1,\dots,d\}$. The off-diagonal entries are $0$.
\vspace{-0.4cm}
\begin{figure}[H]
\begin{center}
\includegraphics[scale=0.55]{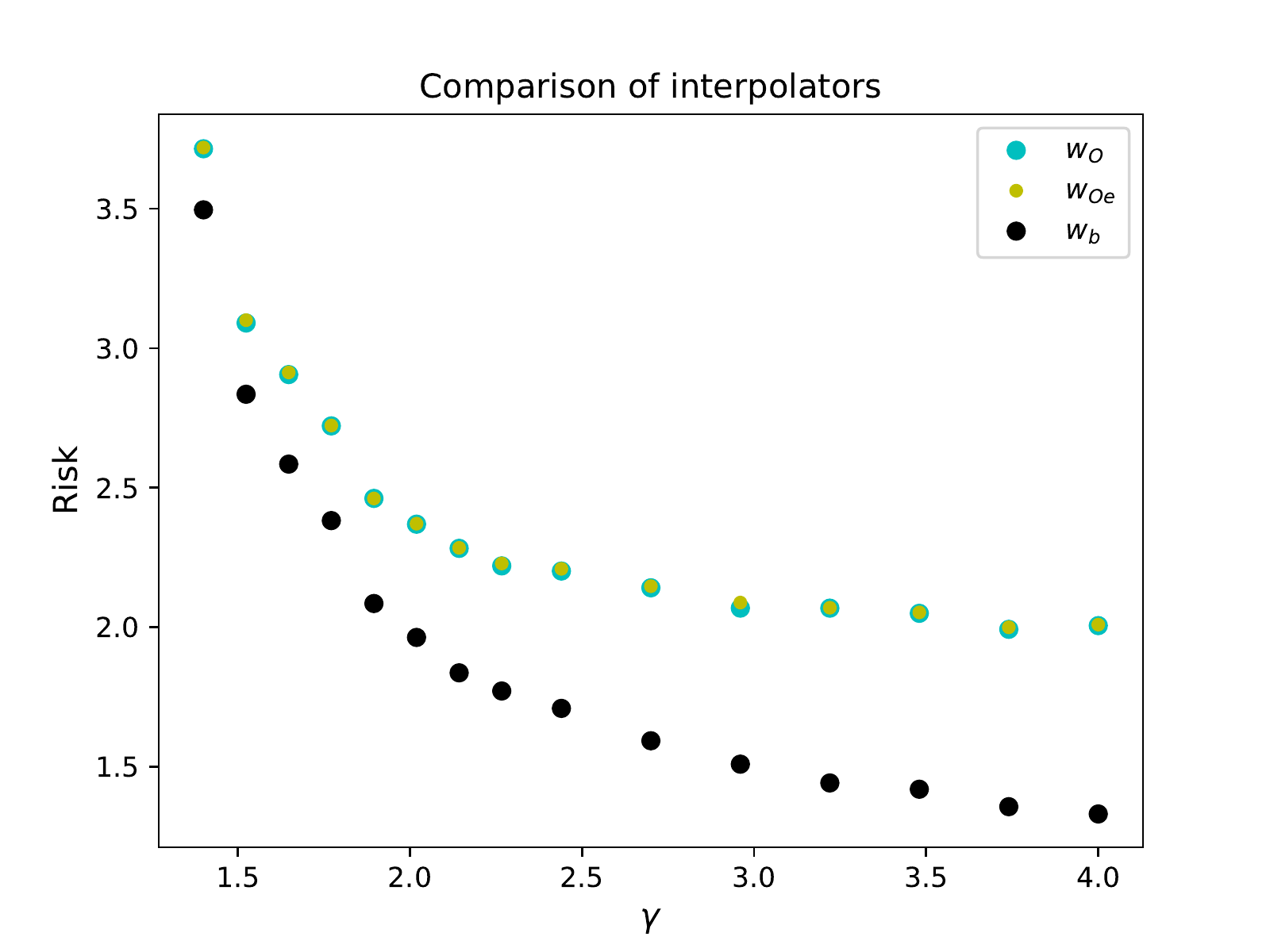}
\caption{Plot of $\E_{\xi}r(w)$ (points) for $w\in\{w_{O}, w_{Oe}, w_b\}$ in the exponential regime with $d = \lfloor \gamma n \rfloor, r^2 = 1, \sigma^2 = 1, n = 2000$.} 
\label{fig_exponential}
\end{center}
\end{figure}
\vspace{-0.45cm}

Note that the Graphical Lasso works well in the regimes of covariance matrices that we presented, because in these regimes the empirical-covariance-based estimator $w_{Oe}$ is seen to reproduce the behaviour of the population-covariance-based estimator $w_{O}$.\\

However, we do not make the claim that the Graphical Lasso approximation will approximate the population covariance matrix well in general. The covariance matrices considered in this work have a notable sparsity structure, and the Graphical Lasso approximation may not perform well for dense covariance matrices.\\

It is interesting to study which covariance matrix approximators one should use. If we consider $\bfc_e = \bfX^T\bfX/n + \lambda I_d$ for any $\lambda\in\R$, one can check using the singular value decomposition of $\bfX$ that
\[{\bfc_{e}}^{-\frac{1}{2}} (\bfX{\bfc_{e}}^{-\frac{1}{2}})^{\dagger}y = \bfX\dg y, \]
so that the corresponding preconditioned gradient descent converges to the same limit as gradient descent and hence removes the benefit of preconditioning. The last statement is also true when using the Ledoit-Wolf shrinkage covariance approximation \citep{ledoitwolf}.
\end{subsection}

\begin{subsection}{Empirical approximation in non-isotropic regimes}\label{subsection_non_isotropic}
In the examples considered so far, we empirically illustrated that $w_{Oe}$ approximates $w_O$ well. However, the considered examples used an isotropic prior, i.e. $\Phi = I_d$. It is natural to ask whether we are also able to match the generalization performance of $w_O$ when $\Phi \ne I_d$.\\

If we knew $\Phi$, or had some prior information about $\Phi$, then we can incorporate this information into an estimate ${\widehat{\Phi}}$ and use the fully empirical approximation
\begin{align}
w_{Oe{\widehat{\Phi}}}  =  \bigg(\frac{\delta_e}{d}{\widehat{\Phi}} X^T  +  {{\Sigma_e}^{-\frac{1}{2}}}(X{{\Sigma_e}^{-\frac{1}{2}}})^\dagger\bigg)\bigg( I_n  +  \frac{\delta_e}{d}X{\widehat{\Phi}}X^T\bigg)^{-1}  y, \label{eqn_wOePhie_definition}
\end{align}
which is likely to perform better than if we used ${\widehat{\Phi}} = I_d$ as in $w_{Oe}$ (\ref{eqn_wOe_definition}).\\

However, in Figures \ref{fig_nonisoregime_1}, \ref{fig_nonisoregime_2} we empirically illustrate  that the interpolator $w_{Oe}$ has generalization very similar to that of $w_O$ and $w_{Oe\Phi}$ even in regimes when the prior is not isotropic ($\Phi \ne I_d$). Using $w_{Oe}$ corresponds to having no information about the prior, while $w_{Oe\Phi}$ corresponds to having complete information about the covariance matrix of the prior. We see that both $w_{Oe\Phi}$ and $w_{Oe}$ approximate $w_O$ well in terms of generalization performance.\\

In Figure \ref{fig_nonisoregime_1}, we consider a prior where $\Phi$ is in the autoregressive regime. That is 
\[\Phi_{ij} = \rho^{|i-j|}\]
for all $i,j\in\{1,\dots, d\}$ and we set $\rho = 0.5$. The population covariance matrix $\Sigma$ is in the exponential regime \citep{dobriban}, where the eigenvalues of $\Sigma$ are evenly spaced quantiles of the standard exponential distribution. Namely, 
\[\Sigma_{ii} = -\text{log}\big(1 - i/(d+1)\big),\] and the off-diagonal entries are $0$. In Figure \ref{fig_nonisoregime_2} we set $\Sigma$ to be in the autoregressive regime with $\rho = 0.5$ and we consider the ``hard prior'' regime \citep{dominic} where $\Phi = \Sigma^{-1}$.

\vspace{-0.3cm}
\begin{figure}[H]
    \begin{minipage}{0.49\textwidth}
        \centering
        \includegraphics[scale=0.45]{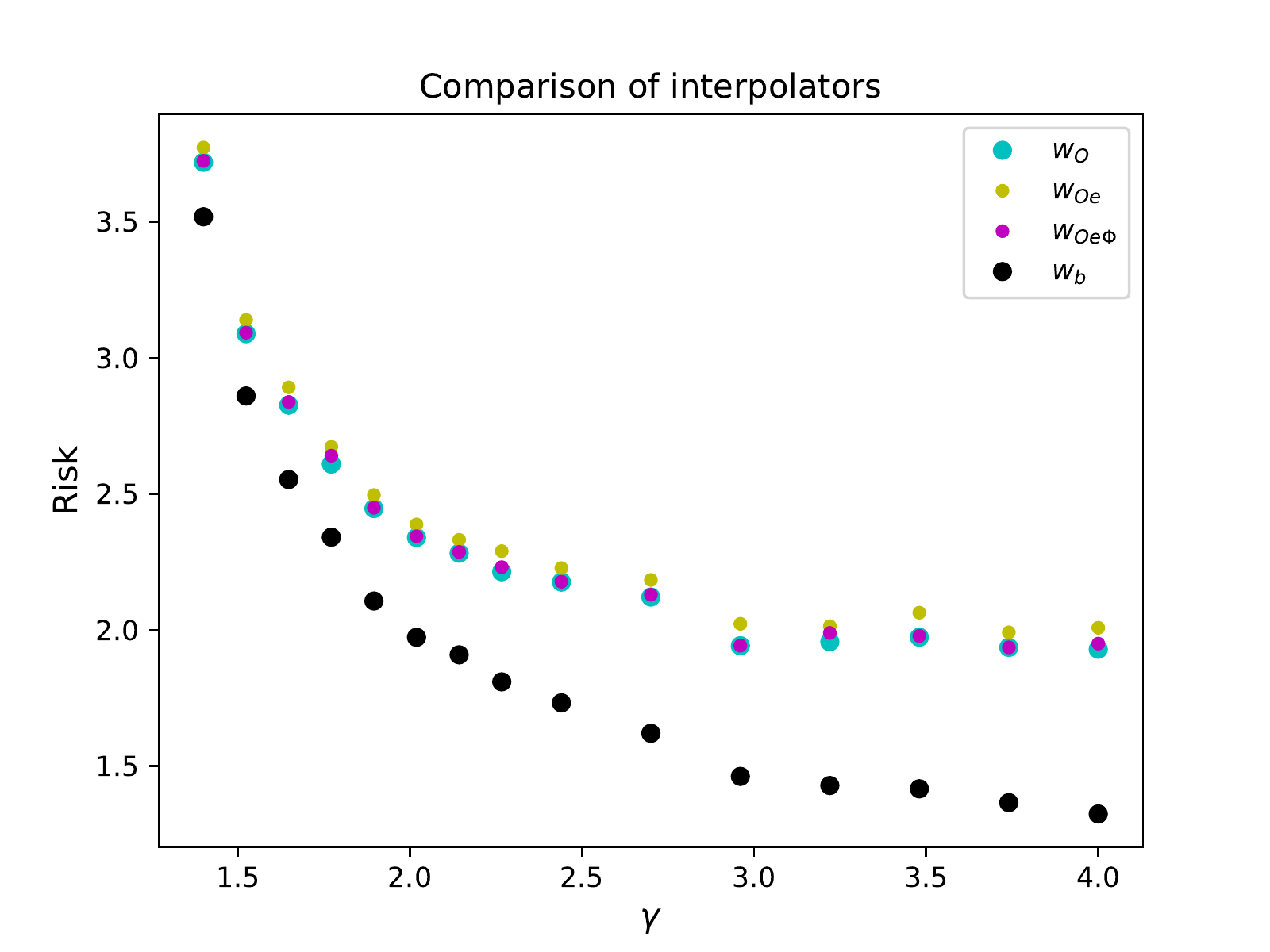}
        \caption{Plot of $\E_{\xi}r(w)$ for $w\in\{w_{O}, w_{Oe}, w_{Oe\Phi}, w_b\}$ with $r^2 = 1, \sigma^2 = 1, \gamma = \lfloor d/n \rfloor, n = 2000$. $\Sigma$ follows the exponential regime and $\Phi$ follows the autoregressive regime with $\rho = 0.5$.} \label{fig_nonisoregime_1}
    \end{minipage}\hspace{0.5em}
    \begin{minipage}{0.49\textwidth}
        \centering
        \includegraphics[scale=0.45]{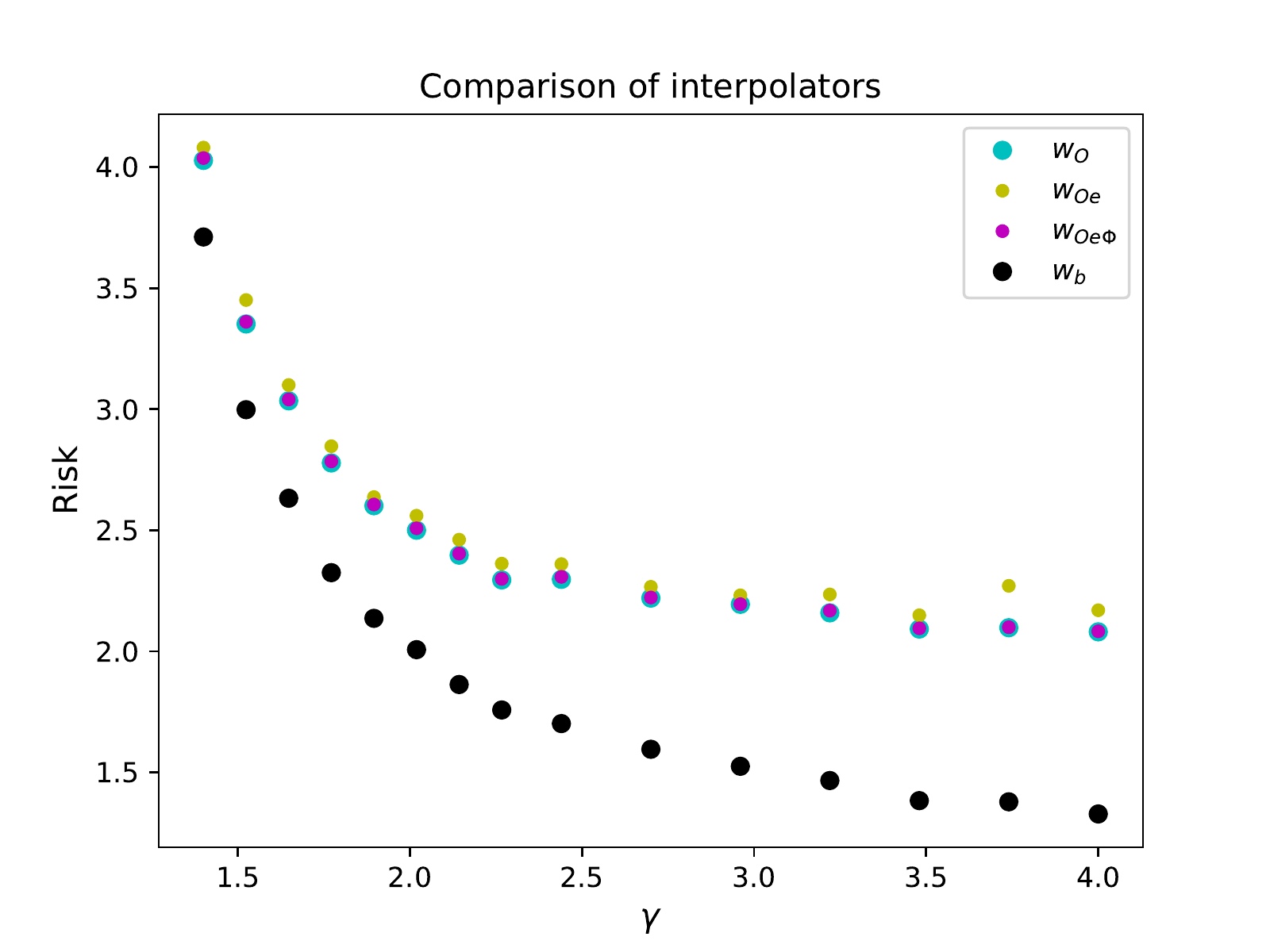}
        \caption{Plot of $\E_{\xi}r(w)$ for $w\in\{w_{O}, w_{Oe}, w_{Oe\Phi}, w_b\}$ with $r^2 = 1, \sigma^2 = 1, \gamma = \lfloor d/n \rfloor, n = 2000$. $\Sigma$ follows the autoregressive regime with $\rho = 0.5$ and $\Phi = \Sigma^{-1}$ follows the ``hard'' prior regime.} \label{fig_nonisoregime_2}
    \end{minipage}
\end{figure}
\vspace{-0.1cm}

\end{subsection}

\begin{subsection}{Proof of Propostion
\ref{prop_bla_rf}}
We prove that the optimal response-linear achievable interpolator in random features regression is
\begin{align*}
a_{O} \!=\! \Sigma_z^{-1}\bigg(\Sigma_{zx}\Phi\bfX^{T}
\!+\! \bfZ^{T}\big(\bfZ\Sigma_z^{-1}\bfZ^{T}\big)^{-1}\big(\frac{d}{\delta}I_n \!+\! \bfX\Phi\bfX^{T} \!-\! \bfZ\Sigma_z^{-1}\Sigma_{zx}\Phi\bfX^{T}\big)\bigg)\bigg(\frac{d}{\delta}I_n \!+\! \bfX\Phi\bfX^T\bigg)^{-1}\hspace{-0.2cm}\!y.
\end{align*}
The proof follows analogous steps to the proof of Proposition \ref{prop_bla}.
\begin{proof}
First, we note that as $a_{O}\in\Lc$, there exists $\R^{d\times n} \ni \bfq = \bfq(\bfX, \bfc, \bfm, \delta)$ such that $a_{O} = \bfq y = \bfq \bfX w^{\star} + \bfq \xi$. Therefore, the definition of $a_{O}$,
\begin{align*}
  a_{O} = \argmin_{a\in\G\cap\Lc} \E_{\xi,w^{\star}} r(f_a) - r(w^{\star}),
\end{align*}
can be restated as
\begin{align}
a_{O} = \argmin_{\substack{a_{O} = \bfq y \\ \bfZ\bfq y = y}}\E_{\xi,w^{\star}} f_1(\bfq) + f_2(\bfq) + f_3(\bfq), \label{eqn_first_restatement_main_prop_rf}
\end{align}
where
\begin{align*}
&f_1(\bfq) = \frac{r^2}{d}\text{Tr}\big(\Sigma_{z}\bfq\bfX\Phi\bfX^T\bfq^T\big)\\
&f_2(\bfq) = \sigma^2\text{Tr}\big(\Sigma_{z}\bfq\bfq^T\big)\\
&f_3(\bfq) = -2\frac{r^2}{d}\text{Tr}\big(\bfq^T\Sigma_{zx}\Phi\bfX^T\big),
\end{align*}
where $\Sigma_z = \E_{\tilde{x}}(\sigma(\Theta \tilde{x}/\sqrt{d})\sigma(\Theta \tilde{x}/\sqrt{d})^T)$ and $\Sigma_{zx} = \E_{\tilde{x}}(\sigma(\Theta \tilde{x}/\sqrt{d})\tilde{x}^T)$. Moreover, $\bfZ\bfq y = y$ almost surely implies that $\bfZ\bfq\bfX = \bfX$ almost surely. This is because taking expectation with respect to $\xi$ in $\bfZ\bfq y = y$ implies
\begin{align*}
    (\bfZ\bfq\bfX - \bfX)w^{\star} = 0
\end{align*}
and therefore
\begin{align*}
    0 = \E_{w^{\star}}\big( \Vert (\bfZ\bfq\bfX - \bfX)w^{\star} \Vert_2^2 \;|\; w^{\star}\big) = {w^{\star}}^T\E\big((\bfZ\bfq\bfX - \bfX)^T(\bfZ\bfq\bfX - \bfX)\big)w^{\star}.
\end{align*}
Because this holds almost surely for all realizations of $w^{\star}\in\Rd$, similarly as in Section \ref{appendix_proof_bla}, it follows that $\E\big((\bfZ\bfq\bfX - \bfX)^T(\bfZ\bfq\bfX - \bfX)\big) = 0$. Finally, therefore also $\E\big(\text{Tr}((\bfZ\bfq\bfX - \bfX)^T(\bfZ\bfq\bfX - \bfX))\big) = 0$ and because $\Vert \bfa \Vert = \sqrt{\text{Tr}(\bfa\bfa^T)}$ is a norm, this implies that $\bfZ\bfq\bfX - \bfX = 0$ almost surely. Hence, (\ref{eqn_first_restatement_main_prop_rf}) is equivalent to 
\begin{align}
w_{O} = \argmin_{\substack{a_{O} = \bfq y \\ \bfZ\bfq \bfX = \bfX}} f_1(\bfq) + f_2(\bfq) + f_3(\bfq). \label{eqn_second_restatement_main_prop_rf}
\end{align}
Now we use Theorem 2 of \cite{penrose_pseudoinv_def} which states that for any matrices $\bfa, \bfb, \bfcc$ and $\bfd$, all solutions $\bfb$ to the equation $\bfa \bfb \bfcc = \bfd$ can be written as $\bfb = \bfa\dg\bfd\bfcc\dg + \bfz - \bfa\dg\bfa\bfz\bfcc\bfcc\dg$ where $\bfz$ is arbitrary. Therefore, $\bfZ\bfq \bfX = \bfX$ is equivalent to
\begin{align*}
\bfq = \bfZ\dg\bfX\bfX\dg + \bfz - \bfZ\dg\bfZ\bfz\bfX\bfX\dg = \bfZ\dg + \bfz - \bfZ\dg\bfZ\bfz,
\end{align*} 
for some arbitrary $\bfz\in\R^{N\times n}$. Hence, (\ref{eqn_second_restatement_main_prop_rf}) is equivalent to an unconstrained optimization problem over $\R^{N\times n}$ in the form
\begin{align}
w_{O} = \argmin_{\bfz}f(\bfz), \label{eqn_third_restatement_main_prop_rf}
\end{align}
where $f(\bfz) = f_1(a_{O}(\bfz)) + f_2(a_{O}(\bfz)) + f_3(a_{O}(\bfz))$. Now we show that $f:\R^{N\times n} \to \R$ is strictly convex. This is done in precisely the same way as in the proof \ref{appendix_proof_bla}. Namely, we note that $\Vert \bfa \Vert = \sqrt{\text{Tr}(\bfa\bfa^T)}$ is a norm and hence $\R^{N\times N}\ni\bfa\mapsto\text{Tr}(\bfa\bfa^T)$ is strictly convex. As $f_1(\bfz),f_2(\bfz)$ are compositions of an affine map with $\R^{N\times N}\ni\bfa\mapsto\text{Tr}(\bfa\bfa^T)$, and by noting that $f_3(\bfz)$ is affine, it follows that $S\mapsto f(S)$ is strictly convex. Moreover, $f$ is differentiable. Therefore, to find a unique global minimum of $f$, it is enough to find $\bfz^{\star}\in\R^{N\times n}$ such that $\partial f(\bfz^{\star}) = 0$. Using tools of matrix calculus we find
\begin{align*}
\partial f(\bfz) = 2(I_N - \bfZ\dg\bfZ)\bfc A,
\end{align*}
where
\begin{align*}
A &= \bigg(\sigma^2\bfz +\frac{r^2}{d}\big(\bfz\bfX - \Sigma_{z}^{-1}\Sigma_{zx}\big)\bfm\bfX^T+\Sigma_{z}\bfZ\dg\big(I_n - \bfZ\bfz\big)\big(\sigma^2I_n +\frac{r^2}{d}\bfX\bfm\bfX^T\big)\bigg).    
\end{align*}
Because $\R^{N} \ni v \mapsto (I_N - \bfZ\dg\bfZ)v$ is the projection onto $\text{Ker}(\bfZ) = \text{Im}(\bfZ^T)^{\bot}$, this hints towards finding $\bfz^{\star}$ such that $A = \bfc^{-1}\bfZ^TB$ for some matrix $B$. This is achieved, for example, if
\begin{align*}
\sigma^2\bfz^{\star} +\frac{r^2}{d}\big(\bfz^{\star}\bfX - \Sigma_{z}^{-1}\Sigma_{zx}\big)\bfm\bfX^T = \bfc^{-1}\bfZ^TB 
\end{align*}
and
\begin{align*}
I_n - \bfZ\bfz^{\star} = 0,
\end{align*}
for some matrix $B$. The first equation implies 
\begin{equation}
\bfz^{\star} = \Sigma_z^{-1}\bigg(\frac{r^2}{d}\Sigma_{zx}\bfm\bfX^T + \bfZ^TB\bigg)\bigg(\sigma^2I_n + \frac{r^2}{d}\bfX\bfm\bfX^T\bigg)^{-1}\label{appendix_auxi_eqn}
\end{equation}
and using that $\bfZ\bfz^{\star} = I_n$ gives
\[B = \bigg(\bfZ\Sigma_{z}^{-1}\bfZ^T\bigg)^{-1}\bigg(\sigma^2I_n + \frac{r^2}{d}\bfX\Phi\bfX^T - \frac{r^2}{d}\bfZ\Sigma_{z}^{-1}\Sigma_{zx}\bfm\bfX^T\bigg).\]
Plugging $B$ back into (\ref{appendix_auxi_eqn}) gives 
\[\bfz^{\star} = \Sigma_z^{-1}\bigg(\Sigma_{zx}\Phi\bfX^{T}
\!+\! \bfZ^{T}\big(\bfZ\Sigma_z^{-1}\bfZ^{T}\big)^{-1}\big(\frac{d}{\delta}I_n \!+\! \bfX\Phi\bfX^{T} \!-\! \bfZ\Sigma_z^{-1}\Sigma_{zx}\Phi\bfX^{T}\big)\bigg)\bigg(\frac{d}{\delta}I_n \!+\! \bfX\Phi\bfX^T\bigg)^{-1}\hspace{-0.2cm}\!.\]
Finaly, because $\bfZ\bfz^{\star} = I_n$, it follows that $\bfq^{\star} = \bfZ\dg + \bfz^{\star} - \bfZ\dg\bfZ\bfz^{\star} = \bfz^{\star}$ and hence
\begin{align*}
a_{O} = \Sigma_z^{-1}\bigg(\Sigma_{zx}\Phi\bfX^{T}
\!+\! \bfZ^{T}\big(\bfZ\Sigma_z^{-1}\bfZ^{T}\big)^{-1}\big(\frac{d}{\delta}I_n \!+\! \bfX\Phi\bfX^{T} \!-\! \bfZ\Sigma_z^{-1}\Sigma_{zx}\Phi\bfX^{T}\big)\bigg)\bigg(\frac{d}{\delta}I_n \!+\! \bfX\Phi\bfX^T\bigg)^{-1}\hspace{-0.2cm}\!y.
\end{align*}
\end{proof}
\end{subsection}

\begin{subsection}{Proof of Proposition \ref{prop_bla_rf_implicit_bias_of_preconditioned_GD}}\label{appendix_subsection_proof_prop_bla_rf}
We prove that the optimal response-linear achievable interpolator $a_{O}$ in random features regression is the limit of preconditioned gradient descent on the last layer,
\begin{align*}
w_{t+1} = w_{t} - \eta_t{\Sigma_{z}}^{-1}\nabla R(w_{t}),
\end{align*}
provided that the algorithm converges and initialized at
\begin{align*}
a_0 = \Sigma_{z}^{-1}\Sigma_{zx}\Phi\bfX^{T}\bigg(\frac{d}{\delta}I_n + \bfX\Phi\bfX^{T}\bigg)^{-1}y.
\end{align*}
\begin{proof}
As before, by a result of \citep{optimization_geometry} we have that the limit of
\begin{align*}
w_{t+1} = w_{t} - \eta_t{\Sigma_{z}}^{-1}\nabla R(w_{t}),
\end{align*}
on the last layer, initialized at some $a_0$ and provided that it converges, satisfies
\begin{align*}
\lim_{t\to\infty}w_{t} = \Sigma_{z}^{-\frac{1}{2}} \big(\bfZ\Sigma_{z}^{-\frac{1}{2}}\big)\dg \big(y - \bfZ a_0\big) + a_0. 
\end{align*}
Using
\begin{align*}
a_0 = \Sigma_{z}^{-1}\Sigma_{zx}\Phi\bfX^{T}\bigg(\frac{d}{\delta}I_n + \bfX\Phi\bfX^{T}\bigg)^{-1}y,
\end{align*}
we obtain
\begin{align*}
&\Sigma_{z}^{-\frac{1}{2}} \big(\bfZ\Sigma_{z}^{-\frac{1}{2}}\big)\dg \big(y - \bfZ a_0\big) + a_0 =\\ 
&\Sigma_z^{-1}\bigg(\Sigma_{zx}\Phi\bfX^{T}
\!+\! \bfZ^{T}\big(\bfZ\Sigma_z^{-1}\bfZ^{T}\big)^{-1}\big(\frac{d}{\delta}I_n \!+\! \bfX\Phi\bfX^{T} \!-\! \bfZ\Sigma_z^{-1}\Sigma_{zx}\Phi\bfX^{T}\big)\bigg)\bigg(\frac{d}{\delta}I_n \!+\! \bfX\Phi\bfX^T\bigg)^{-1}\hspace{-0.2cm}\!y. 
\end{align*}
\end{proof}
\end{subsection}
\begin{subsection}{Random features example}\label{appendix_section_rf_example}
We illustrate the test error of the random features model $x\mapsto f_a(x) = a^T\sigma(\Theta x/\sqrt{d})$ for the optimal response-linear achievable interpolator, $f_{a_{O}}$, with
\begin{align*}
a_{O} \!=\! \Sigma_z^{-1}\bigg(\Sigma_{zx}\Phi\bfX^{T}
\!+\! \bfZ^{T}\big(\bfZ\Sigma_z^{-1}\bfZ^{T}\big)^{-1}\big(\frac{d}{\delta}I_n \!+\! \bfX\Phi\bfX^{T} \!-\! \bfZ\Sigma_z^{-1}\Sigma_{zx}\Phi\bfX^{T}\big)\bigg)\bigg(\frac{d}{\delta}I_n \!+\! \bfX\Phi\bfX^T\bigg)^{-1}\hspace{-0.2cm}\!y
\end{align*}
in comparison to the test error for the minimum-norm interpolator $a_{\ell_2} = \bfZ\dg y$ on a standard example. Let $x_i\sim\textrm{Unif}(\mathbb{S}^{d-1}(\sqrt{d}))$ and $\Theta\in\R^{N\times d}$ be randomly initialized such that the rows of $\Theta$ satisfy $\Theta_i\in\mathbb{S}^{d-1}(\sqrt{d})$. Here, $\mathbb{S}^{d-1}(\sqrt{d})$ is the sphere with radius $\sqrt{d}$ in $\Rd$. We numerically compute $\Sigma_z = \E_{\tilde{x}}(\sigma(\Theta \tilde{x}/\sqrt{d})\sigma(\Theta \tilde{x}/\sqrt{d})^T)$ and $\Sigma_{zx} = \E_{\tilde{x}}(\sigma(\Theta \tilde{x}/\sqrt{d})\tilde{x}^T)$ by sampling from $\textrm{Unif}(\mathbb{S}^{d-1}(\sqrt{d}))$ and use the true signal-to-noise ratio $\delta$. We observe, as expected, that $f_{a_O}$ generalizes better than $f_{a_{\ell_2}}$. This is so even for large $\gamma = \lfloor N/d \rfloor$, where \citet{mei2019generalization} showed that, for high-enough signal-to-noise ratio, the test error of the minimum-norm interpolator converges to the test error of the optimally-tuned ridge regression estimator in the limit as $N/d\to\infty$ (under certain assumptions).
\vspace{-0.3cm}
\begin{figure}[H]
\centering
\includegraphics[scale=0.55]{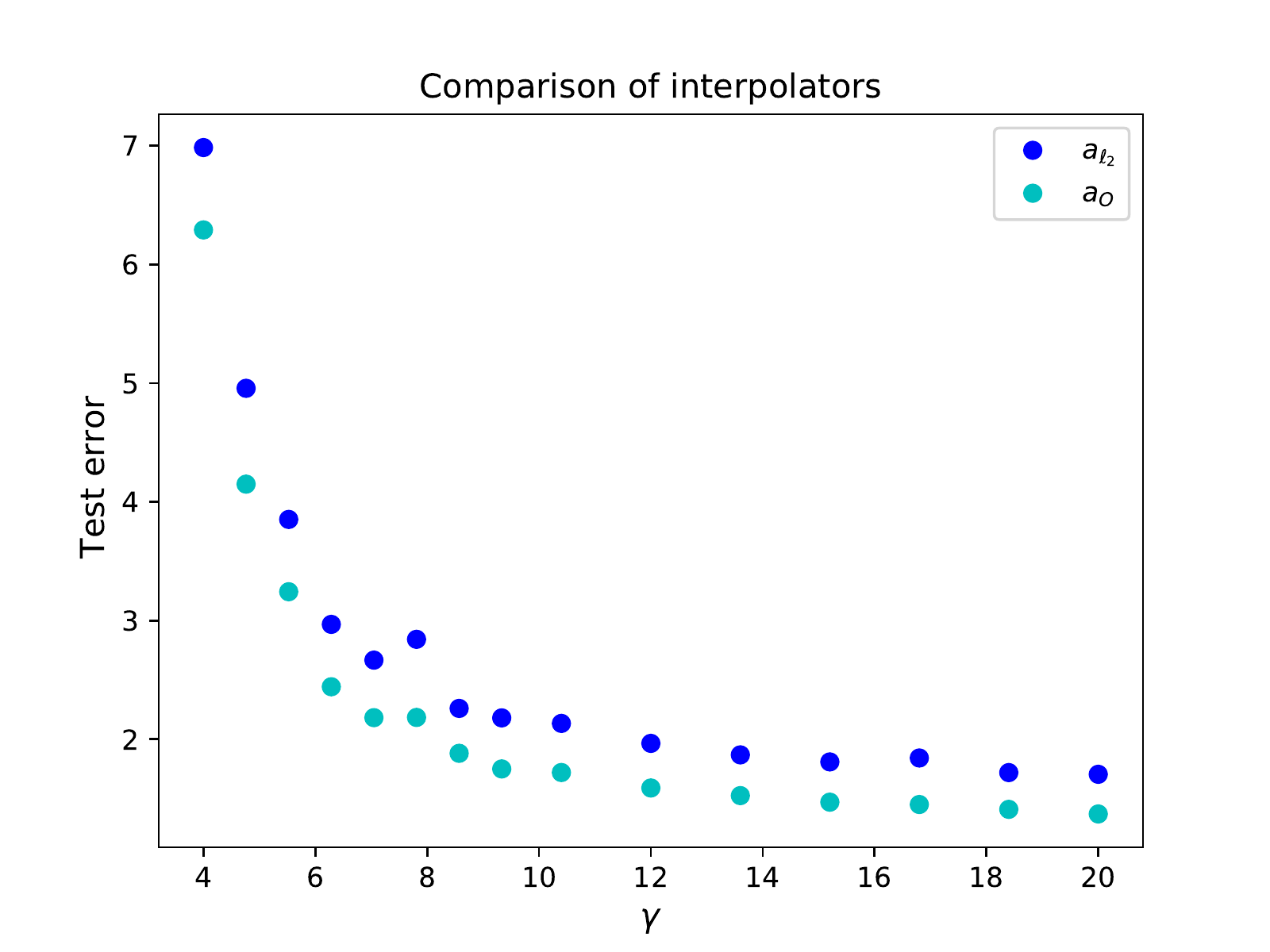}
\caption{Plot of test error of $f\in\{f_{a_{\ell_2}}, f_{a_{O}}\}$ for $\gamma = \lfloor N/d \rfloor$ when $x_i\sim\textrm{Unif}(\mathbb{S}^{d-1}(\sqrt{d}))$ with $r^2 = 5, \sigma^2 = 1, \lfloor n/d\rfloor = 3, n = 2000$.} \label{fig_rf}
\end{figure}
\vspace{-0.45cm}
\end{subsection}

\end{document}